\documentclass[10pt,twocolumn,letterpaper]{article}

\usepackage[pagenumbers]{cvpr}

\usepackage{todonotes}
\reversemarginpar

\usepackage{xspace}
\newcommand{\model}{NEPA\xspace}

\usepackage{booktabs}
\usepackage{multirow}
\usepackage{graphicx}
\usepackage{algorithm}
\usepackage{algpseudocode}
\usepackage{xcolor}
\usepackage{listings}
\usepackage{pifont}
\usepackage{tabularx}
\usepackage[table]{xcolor}
\newcommand{\cmark}{\checkmark}%
\newcommand{\xmark}{\ding{55}}%
\usepackage{array,subcaption,siunitx}

\definecolor{lightgray}{gray}{0.92}
\newcolumntype{Y}{>{\centering\arraybackslash}X}
\renewcommand{\arraystretch}{1.15}

\definecolor{cvprblue}{rgb}{0.21,0.49,0.74}
\usepackage[pagebackref,breaklinks,colorlinks,allcolors=cvprblue]{hyperref}

\def\paperID{} 
\def\confName{CVPR}
\def\confYear{2026}

\title{Next-Embedding Prediction Makes Strong Vision Learners}

\newcommand{\umich}{$^{1}$}
\newcommand{\nyu}{$^{2}$}
\newcommand{\pton}{$^{3}$}
\newcommand{\uva}{$^{4}$}
\newcommand{\visiting}{$^{\ddag}$}

\author{%
    Sihan Xu\umich\hspace{10pt} 
    Ziqiao Ma\umich\hspace{10pt} 
    Wenhao Chai\pton\hspace{10pt}
    Xuweiyi Chen\uva\hspace{10pt}
    Weiyang Jin\visiting\nyu\hspace{10pt} \\
    Joyce Chai\umich\hspace{10pt}
    Saining Xie\nyu\hspace{10pt}
    Stella X. Yu\umich\hspace{10pt}  \\
    \umich University of Michigan \quad
    \nyu New York University \quad
    \pton Princeton University \quad
    \uva University of Virginia \\
    \texttt{\url{https://sihanxu.me/nepa}}
}

\begin{document}
\maketitle

\def\thefootnote{$^{\ddag}$}\footnotetext{Work partly done when visiting NYU.}

\begin{abstract}

Inspired by the success of generative pretraining in natural language, we ask whether the same principles can yield strong self-supervised visual learners.
Instead of training models to output features for downstream use, we train them to generate embeddings to perform predictive tasks directly. 
This work explores such a shift from learning representations to learning models.
Specifically, models learn to predict future patch embeddings conditioned on past ones, using causal masking and stop gradient, which we refer to as Next-Embedding Predictive Autoregression (\model).
We demonstrate that a simple Transformer pretrained on ImageNet-1k with \mbox{next embedding prediction} as its sole learning objective is effective---no pixel reconstruction, discrete tokens, contrastive loss, or task-specific heads. 
This formulation retains architectural simplicity and scalability, without requiring additional design complexity.
\model achieves strong results across tasks, attaining 83.8\% and 85.3\% top-1 accuracy on ImageNet-1K with ViT-B and ViT-L backbones after fine-tuning, and transferring effectively to semantic segmentation on ADE20K.
We believe generative pretraining from embeddings provides a simple, scalable, and potentially modality-agnostic alternative to visual self-supervised learning.

\end{abstract}    
\section{Introduction}
\label{sec:intro}

Visual pretraining is one of the core topic in computer vision.
Self-supervised learning has become the cornerstone of modern visual pretraining method, enabling scalable visual learners without manual labels. 
At its core, the objective is to learn representations: models are optimized to map raw pixels to a fixed-dimensional representation, which can later be used or fine-tuned for downstream tasks.
This philosophy unites methods based on instance discrimination~\citep{chen2020simclr, he2020moco}, self-distillation~\citep{caron2021dino}, and masked reconstruction~\citep{he2022mae, bao2022beit}.
The goal is to learn visual representations that can be consumed by downstream modules at various scales, from lightweight task-specific heads to large cascaded systems such as vision-language models.

\begin{figure}[t]
    \centering
    \includegraphics[width=0.7\linewidth]{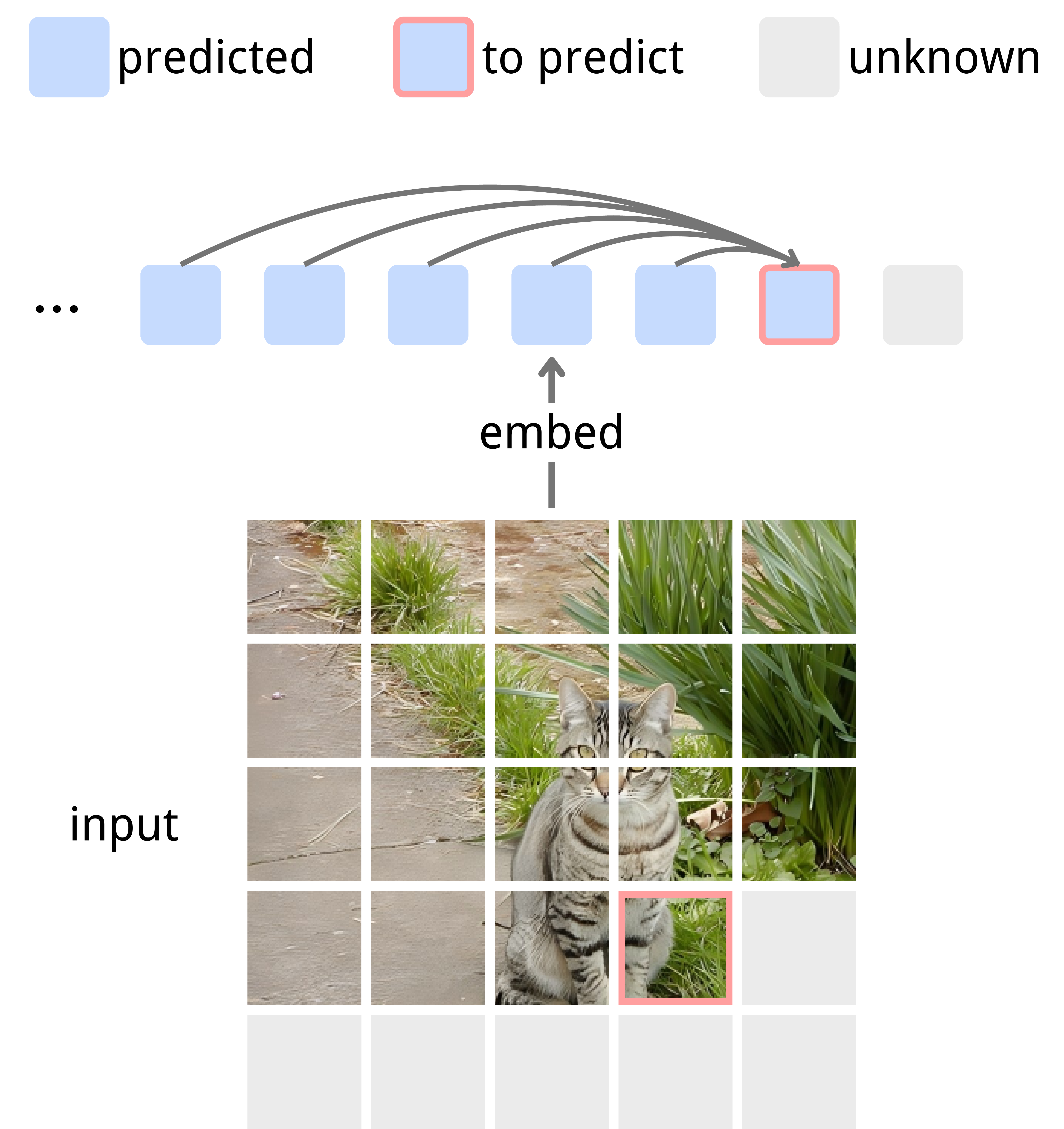}
    \caption{Next-Embedding Predictive Autoregression (\model). An image is split into patches and embedded into a sequence. An autoregressive model predicts the next embedding from previous ones, mirroring next-token prediction in language models.}
    \vspace{-10pt}
    \label{fig:teaser}
\end{figure}

A fundamentally different paradigm has underpinned the success of modern natural language processing. 
Language models are not pretrained to be feature extractors; they are trained to be \textit{generative and predictive systems}. 
The objective is not to produce a static embedding of a sentence, but to model the data distribution itself through a simple, causal objective~\citep{radford2018gpt, radford2019gpt2, brown2020gpt3, achiam2023gpt4}. 
This training compels the model to internalize the semantics and conditional dependencies within language. 
Inference is no longer a two-stage process of encoding followed by task-solving, but a single predictive computation carried out by the model itself. 
This distinction is fundamental.
It suggests that \textit{generative prediction}, rather than \textit{representation learning}, may offer a direct route to scale up pretraining. 
A very recent line of research has shifted to this philosophy, \eg, early pixel-level generative pretraining (iGPT) showed transferable features but struggled with very long sequences and weak semantic alignment~\citep{chen2020igpt}.
JEPA~\citep{assran2023jepa} moves beyond pixels by predicting latent targets and aligning more closely with semantic structure. 
Still, JEPA trains by regressing to latent targets from a momentum encoder, rather than using generative prediction as the self-supervised objective.

We are thus tempted to ask whether minimal causal pretraining might also give rise to strong vision learners.
Concretely, an image is decomposed into patches, which are mapped into a sequence of patch-wise embeddings. 
A causal transformer is then trained to predict the next embedding given all previous ones, closely mirroring the next-token prediction paradigm in language models. 
We use a stop-gradient on the target embeddings to create a stable predictive task.
This formulation is deliberately minimalist. 
It requires no pixel-level decoder, no discrete visual tokenizer, and none of the engineered data augmentations, negative pairs, or momentum encoders common in contrastive learning. 
The entire learning signal is derived from the model's ability to anticipate the future in embedding space. 
We call this new family of models \textit{Next-Embedding Predictive Autoregression (\model)}.

The effectiveness of this simplicity is demonstrated through its performance. 
A standard Vision Transformer, pretrained on ImageNet-1K using only our next-embedding prediction objective, achieves state-of-the-art classification accuracy after standard fine-tuning. 
Furthermore, the learned model transfers effectively to dense prediction tasks, achieving strong results on ADE20K semantic segmentation. 
These results validate that a purely predictive model, free from the architectural complexity of other paradigms, can learn the rich, contextual features required for diverse visual tasks.

\vspace{-5pt}
\section{Related Work}
\label{sec:background}

\paragraph{Visual Self-Supervised Learning.}
Self-supervised learning has become the leading paradigm for visual learning, leveraging large unlabeled datasets to learn representations without annotations~\citep{wang2015video,noroozi2016puzzles,zhang2016color,pathak2017move,gidaris2018rotation}. 
Early progress was driven by contrastive and self-distillation approaches~\citep{hadsell2006dimensionality,wu2018unsupervised,oord2018cpc,hjelm2018dim,bromley1993siamese,he2020moco,chen2021mocov3,chen2020simsiam,chen2020simclr,grill2020byol,caron2021dino}, which learn invariances across augmented views but often require large batches or memory banks. 
A parallel line uses reconstruction objectives, where lightweight decoders recover masked pixels or tokens at scale~\citep{vincent2008dae,pathak2016context,devlin019bert,he2022mae,bao2022beit}. 
The limitations of reconstruction have motivated predictive representation learning, which forecasts semantic embeddings rather than raw inputs, as in JEPA~\citep{assran2023jepa}. 
Yet JEPA remains fundamentally representation-centric: its pretrained encoder produces features that downstream modules must consumed with separate heads or fine-tuning, rather than directly driving task behavior through prediction itself.

\vspace{-10pt}
\paragraph{Generative Pretraining.}
Generative pretraining has driven major advances in language~\citep{radford2018gpt, radford2019gpt2, brown2020gpt3, achiam2023gpt4} and has a long visual lineage.
iGPT autoregresses over discrete pixel tokens and learns transferable features, but suffers from long sequences and weak semantic alignment~\citep{chen2020igpt}.
AIM series~\cite{el2024scalable,fini2025multimodal} extend to the continuous counterpart, supervised on predefined pixel-space targets as continuous patch values.
Subsequent work introduced discrete tokenizers~\citep{van2017vqvae, esser2021vqgan, lee2022rqvae, sun2024llamagen, pang2024randar} to compress images into short code sequences amenable to transformer generative pretraining~\citep{esser2021vqgan, ramesh2021dalle, lee2022rqvae}, enabling large-scale image generation. 
Recent scaling efforts, such as VAR~\citep{tian2024var} and LlamaGen~\citep{sun2024llamagen}, further close or surpass the diffusion modeling performance. 
In parallel, tokenizer-free generative pretraining operates directly in continuous spaces~\citep{fan2024fluid, li2024mar}, and masked autoregression blends bidirectional context with ordered generation for efficiency and quality~\citep{li2024mar, wu2025dcar, chang2022maskgit, li2023mage}.
Unlike these works, we treat prediction itself as the learning signal and operate directly in the embedding space without a generative decoder.

\vspace{-10pt}
\paragraph{Representation Prediction.}
Predicting representations instead of raw signals is a long-standing idea in both neuroscience and machine learning.
Predictive coding suggests that the brain forecasts latent causes of sensory inputs~\citep{rao1999predictive}, while Contrastive Predictive Coding (CPC) and related frameworks apply this principle in practice~\citep{oord2018cpc,wu2018unsupervised,hjelm2018dim,bromley1993siamese,he2020moco,chen2021mocov3,chen2020simsiam,chen2020simclr,grill2020byol,caron2021dino}. 
JEPA makes this explicit by predicting target embeddings from a context view in a shared space~\citep{assran2023jepa}, and CLIP can be interpreted as cross-modal embedding prediction aligning text and image representations~\citep{radford2021clip}. 
However, current JEPA-style systems typically rely on auxiliary encoders and prediction heads and predict in parallel rather than causally, contrasting with the simple, scalable next-token objective that underpins generative pretrained language models, which is a formulation proven both effective and efficient.

\vspace{-5pt}
\section{Method}
\label{sec:method}

In this section, we first introduce the next-embedding prediction objective, followed by the architectural design that enables both pretraining and downstream fine-tuning.

\subsection{Next Embedding Prediction}

Given an input image $x$, we divide it into $T$ non-overlapping patches and map each patch to an embedding using a shared encoder $f$, resulting in a sequence of embeddings $z = \{z_1, z_2, \ldots, z_T\}$.
We consider \textit{next embedding prediction} as the training objective, where an autoregressive predictor $h_\theta$ aims to model the next embedding conditioned on the previous ones:
\begin{equation}
\widehat{z}_{t+1} = h_\theta(z_{\leq t}),
\end{equation}
This is directly analogous to next-token prediction in language modeling, but operates entirely in continuous embedding space rather than over discrete tokens or pixel values.

To optimize this objective, we adopt a similarity-based loss inspired by SimSiam~\citep{chen2020simsiam}. 
At each position $t$, the model predicts the next embedding $\widehat{z}_{t+1}$ based on the context $z_{\leq t}$, 
and compares it to the corresponding embedding $z_{t+1}$ obtained directly from the encoder. 
We treat $\widehat{z}_{t+1}$ as the \textit{query} and $z_{t+1}$ as the \textit{target}.
Both vectors are normalized to unit length, and the similarity is measured via negative cosine similarity:
\begin{equation}
\mathcal{D}(z, \widehat{z}) = - \frac{1}{T-1} \sum_{t=1}^{T-1} 
(\frac{z_{t+1}}{\|z_{t+1}\|_2}
\cdot
\frac{\widehat{z}_{t+1}}{\|\widehat{z}_{t+1}\|_2}).
\end{equation}
To avoid degenerate solutions, we follow~\citet{chen2020simsiam} and apply a stop-gradient operation to the target embeddings.
The final training loss is defined as:
\begin{equation}
\mathcal{L} = \mathcal{D}(\operatorname{stopgrad}(z), \widehat{z}).
\end{equation}

Intuitively, this objective encourages the model to predict embeddings that are semantically aligned with the ground truth, without requiring explicit reconstruction of the input. 
The training can be easily implemented using Algorithm~\ref{alg:code}.

\begin{algorithm}[!t]
\caption{Next-Embedding Prediction}
\label{alg:code}
\definecolor{codeblue}{rgb}{0.25,0.5,0.5}
\definecolor{codekw}{rgb}{0.85, 0.18, 0.50}
\lstset{
  backgroundcolor=\color{white},
  basicstyle=\fontsize{7.5pt}{7.5pt}\ttfamily\selectfont,
  columns=fullflexible,
  breaklines=true,
  captionpos=b,
  commentstyle=\fontsize{7.5pt}{7.5pt}\color{codeblue},
  keywordstyle=\fontsize{7.5pt}{7.5pt}\color{codekw},
  escapeinside={@@}{@@}, keepspaces=true,
}

\begin{lstlisting}[language=python]
# f: embedding layer
# h: autoregressive model

for pixel_values in loader: # x, [B, H, W, C]
    input_embed = f(pixel_values) # z, [B, T, D]
    pred_embed  = h(input_embed) # z_hat, [B, T, D]

    loss = D(input_embed, pred_embed) # loss

    loss.backward() # back-propagate
    update(f.param, h.param) # update parameters

def D(z, z_hat):
    target = z.detach() # stop gradient

    pred   = z_hat[:, 0:T-1, :] # shift, [B, T-1, D]
    target = target[:, 1:T, :] # shift, [B, T-1, D]

    # Use any suitable distance metric.
    pred   = normalize(pred, axis=-1) # l2-norm
    target = normalize(target, axis=-1) # l2-norm
    return -(pred * target).sum(dim=-1).mean()
\end{lstlisting}
\end{algorithm}

\subsection{Model Architecture}
\label{sec:model}

\begin{figure}[t]
    \centering
    \includegraphics[width=1.0\linewidth]{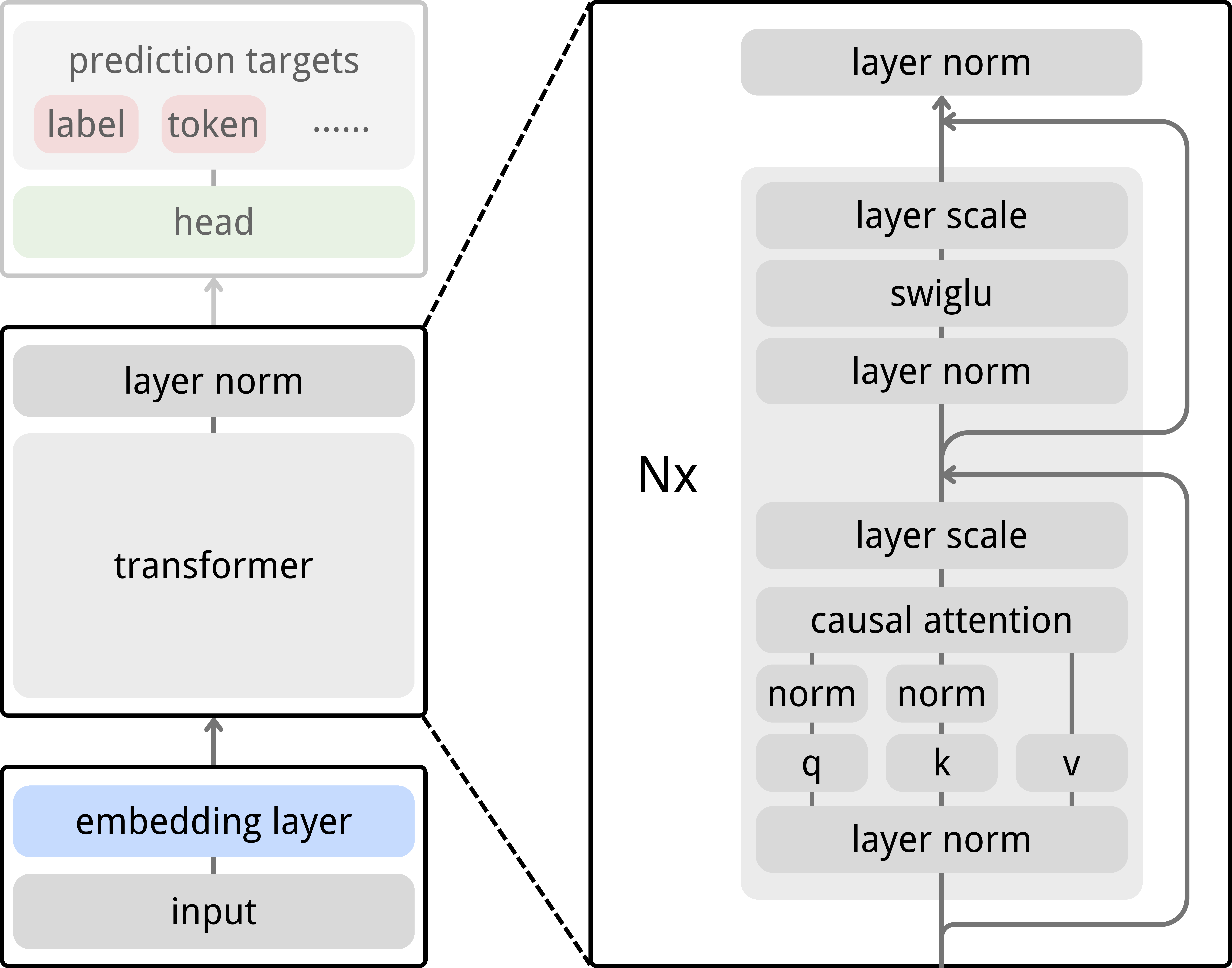}
    \caption{Images are tokenized via a Conv2d patch embedder before entering a pre-norm Transformer with LayerNorm. Modern stabilization components (RoPE~\citep{su2024roformer}, LayerScale~\citep{touvron2021cait}, SwiGLU~\citep{shazeer2020glu}, and QK-Norm~\citep{henry2020qknorm}) are applied at all layers. \vspace{-5pt}}
    \label{fig:model}
\end{figure}

We adopt a standard Vision Transformer (ViT) backbone~\citep{dosovitskiy2020vit} with causal attention masking. 
Unlike pixel-level reconstruction methods~\citep{he2022mae, bao2022beit}, our approach requires no separate decoder. 
The Transformer predicts future patch embeddings directly from past ones, using a single backbone for both context encoding and prediction, similar to autoregressive language models.
Images are split into non-overlapping patches using a Conv2d patch embedding layer, with learnable positional embeddings added before being fed into the Transformer. 
We adopt a pre-norm design with LayerNorm~\citep{ba2016layernormalization} and apply a final LayerNorm to the output features. 
To improve stability and scalability, we incorporate modern training and normalization practices inspired by DINOv3~\citep{siméoni2025dinov3} and VisionLLaMA~\citep{chu2024visionllama}, as shown in Figure~\ref{fig:model}.
These modeling designs are helpful for training but orthogonal to our core framework, and are included below for reproducibility and completeness.

\vspace{-10pt}
\paragraph{RoPE.} 
We adopt Rotary Position Embedding (RoPE~\citep{su2024roformer}) at all layers to encode relative positions via complex rotations in attention. 
RoPE improves generalization and positional reasoning over varying sequence lengths~\citep{touvron2023llama,bai2023qwentechnicalreport,siméoni2025dinov3}.

\vspace{-10pt}
\paragraph{LayerScale.} 
We adopt LayerScale~\citep{touvron2021cait} to stabilize training by applying learnable per-channel scales (initialized to $10^{-5}$) to residual branches, thereby improving convergence with minimal computational overhead.

\vspace{-10pt}
\paragraph{SwiGLU.} 
We replace the standard GeLU activation~\citep{hendrycks2017gelu} in vision transformer feed-forward networks~\citep{dosovitskiy2020vit} with the SwiGLU activation~\citep{shazeer2020glu}. 
While our experiments (Section~\ref{sec:exp}) show only modest improvements over GeLU, we retain SwiGLU to align with recent architectures~\citep{oquab2023dinov2, siméoni2025dinov3, chu2024visionllama, sun2024llamagen} and to ensure compatibility with prevailing designs in both advanced vision models and large language models~\citep{touvron2023llama, bai2023qwentechnicalreport}.

\vspace{-10pt}
\paragraph{QK-Norm.} 
To further enhance training stability, we adopt query-key normalization (QK-Norm~\citep{henry2020qknorm}). This helps mitigate issues such as gradient explosion or collapse in attention and facilitates optimization in deeper transformers~\citep{dehghani2023vit22b, esser2024sd3}. 
In our implementation, we apply LayerNorm~\citep{ba2016layernormalization} without bias and without learnable parameters, ensuring a lightweight yet effective normalization scheme.

\subsection{Task-Specific Heads}

Vision tasks such as image classification, semantic segmentation, object detection, and generative modeling differ in output structure and supervision signals. 
Among visual tasks, image classification and semantic segmentation are the most commonly used benchmarks for evaluating pretrained models, as they reflect global and dense prediction capabilities.
Thus, in our evaluation, we focus primarily on these two tasks.
To evaluate the downstream utility of our pretrained model, we design task-specific heads for classification and segmentation, respectively. 
These heads are lightweight, compatible with standard practices, and do not modify the model architecture.

\vspace{-10pt}
\paragraph{Classification Head.} 
For image-level recognition tasks, we adopt a standard linear classification head~\citep{dosovitskiy2020vit, bao2022beit, he2022mae, caron2021dino, assran2023jepa}.
Specifically, given the final output hidden states $\{z_1, z_2, \ldots, z_T\}$ from the pretrained transformer, we select the last embedding $z_T$ as a compact representation and apply a linear classifier to predict the category label. 
We train this head using the cross-entropy loss over the predicted logits $\widehat{y}$ and the ground-truth label $y$.

\vspace{-10pt}
\paragraph{Segmentation Head.}  
Following previous work~\citep{he2022mae, bao2022beit}, for semantic segmentation, we adopt a UPerNet head~\citep{xiao2018upernet} that operates on multi-scale features extracted from intermediate transformer blocks. 
We supervise predictions using the standard pixel-wise cross-entropy loss. 
To enable spatial context modeling, we disable causal masking and allow bidirectional attention during fine-tuning.

\section{Experiments}
\label{sec:exp}


\begin{table*}[t]
\centering
\scriptsize

\begin{minipage}[t]{0.49\linewidth}
\vspace{0pt}
\centering
\begin{tabular}{c c c c}
\toprule
\textbf{shifting} & \textbf{causal masking} & \textbf{stop-grad} & {\textbf{50k acc}~(\%)} \\
\midrule
\xmark & \cmark & \cmark & {\color{red}{fail}} \\
\cmark & \xmark & \cmark & 73.6 \\
\cmark & \cmark & \xmark & {\color{red}{fail}} \\
\rowcolor{lightgray}
\cmark & \cmark & \cmark & \bfseries 76.8 \\
\bottomrule
\end{tabular}
\subcaption{Effects of causal masking, autoregressive shifting, and stop-gradient. Removing shifting leads to divergence during fine-tuning, while removing stop-gradient results in training collapse during pretraining.}
\label{tab:algo_desing}
\end{minipage}
\hfill
\begin{minipage}[t]{0.49\linewidth}
\vspace{0pt}
\centering
\begin{tabular}{c c c c c}
\toprule
\textbf{LayerScale} & \textbf{RoPE} & \textbf{QK-Norm} & \textbf{GatedMLP} & {\textbf{100k acc}~(\%)} \\
\midrule
\xmark & \xmark & \xmark & \xmark & 78.2 \\
\cmark & \xmark & \xmark & \xmark & 77.4 \\
\cmark & \cmark & \xmark & \xmark & 80.2 \\
\cmark & \cmark & \xmark & \cmark & {\color{red}{fail}} \\
\cmark & \cmark & \cmark & \xmark & 81.1 \\
\rowcolor{lightgray}
\cmark & \cmark & \cmark & \cmark & \bfseries 81.3 \\
\bottomrule
\end{tabular}
\subcaption{Ablation of architectural components on 100k-step accuracy.}
\label{tab:model_desing}
\end{minipage}

\vspace{0.6em}

\begin{minipage}[t]{0.32\linewidth}
\vspace{0pt}
\centering
\begin{tabular}{c c}
\toprule
\textbf{masking ratio} & {\textbf{100k acc}~(\%)}\\
\midrule
\rowcolor{lightgray}
0  & \bfseries 78.2 \\
40 & 76.4 \\
60 & 75.7 \\
\bottomrule
\end{tabular}
\subcaption{Effect of input embedding masking ratio on 100k-step pretraining accuracy using a baseline ViT backbone with embedding prediction.}
\label{tab:masking}
\end{minipage}
\hfill
\begin{minipage}[t]{0.32\linewidth}
\vspace{0pt}
\centering
\begin{tabular}{c c}
\toprule
\textbf{freeze embedding}  & {\textbf{100k acc}~(\%)} \\
\midrule
\xmark & 73.1 \\
\rowcolor{lightgray}
\cmark & \bfseries 77.4 \\
\bottomrule
\end{tabular}
\vspace{10pt}
\subcaption{Effect of freezing the patch embedding layer in the base model with LayerScale on 100k-step pretraining accuracy.}
\label{tab:embed}
\end{minipage}
\hfill
\begin{minipage}[t]{0.32\linewidth}
\vspace{0pt}
\centering
\begin{tabular}{c c}
\toprule
\textbf{attn type} & {\textbf{100k acc}~(\%)} \\
\midrule
\rowcolor{lightgray}
bidirect  & \bfseries 82.5 \\
causal & 81.3 \\
\bottomrule
\end{tabular}
\vspace{10pt}
\subcaption{Effect of attention type during fine-tuning for our full model. We compare causal and bidirectional attention on 100k-step pretraining accuracy.}
\label{tab:attn}
\end{minipage}

\caption{\textbf{Ablations studies.} Default settings are marked in \colorbox{lightgray}{gray}. Experiments that fail to converge are marked with {\color{red}{fail}}.}
\label{tab:ablation}
\vspace{-5pt}
\end{table*}

We conduct comprehensive experiments to evaluate \model of different sizes and configurations. 
All models are pretrained from scratch on the ImageNet-1K dataset~\citep{russakovsky2015imagenet1k} without labels.
We begin with a controlled study on the contributions of key components in our next embedding prediction algorithm, including causal masking, temporal shifting, and stop-gradient. 
We then ablate a series of architectural design choices that further improve performance. 
In comparison to previous work, we evaluate \model on two standard benchmarks: ImageNet-1K~\citep{russakovsky2015imagenet1k} for image classification and ADE20K~\citep{zhou2017ade20k} for semantic segmentation.
To better understand the model's internal behavior, we examine the learned attention patterns and embedding. 

\subsection{Ablation Study on Core Algorithm Design}

This section separately studies the impact of masking, autoregressive shifting, and stop-gradient, which are core algorithm designs for the next embedding prediction objective.
All models are pretrained with a batch size of 2048 for either 50k or 100k steps, and then fine-tuned for downstream tasks; full training details are provided in Appendix~\ref{app:training-details}. 
All reported results are obtained using the exponential moving average (EMA~\citep{tarvainen2017mean}) model with an EMA decay rate of $0.9999$.
We report the top-1 accuracy on the ImageNet-1K validation set to compare different settings.
Tables~\ref{tab:algo_desing} and~\ref{tab:masking} outline the results.

\vspace{-10pt}

\paragraph{Shifting.} 
In our default setup, the model predicts the embedding of the next token rather than copying the current input. 
Removing this shift reduces the objective to an identity mapping, which provides no meaningful prediction target. 
After 50k steps of pretraining, this variant diverges during fine-tuning and fails to converge to a usable model.

\vspace{-10pt}

\paragraph{Causal masking.} 
By default, each token attends only to its predecessors, ensuring a causal prediction setup. 
Removing the causal mask allows every token to attend bidirectionally, effectively turning the model into a reconstruction model. 
After 50k steps of pretraining, this variant achieves only 73.6\% top-1 accuracy after fine-tuning, compared to 76.8\% with causal masking.

\vspace{-10pt}

\paragraph{Stop-gradient.} 
Following SimSiam~\citep{chen2020simsiam}, we stop gradients on the ground-truth embeddings to prevent collapse. 
Without this operation, the training loss collapses to \(-1\) during pretraining, indicating all embeddings are the same.

\vspace{-10pt}

\paragraph{Random masking.} 
We also applied random masking to the input embeddings, while still predicting all target embeddings. 
In Table~\ref{tab:masking}, we tested different masking ratios at 100k pretraining steps and observed a clear performance drop as masking increased: 0\% masking achieved 78.2\% top-1 accuracy, 40\% masking achieved 76.4\%, and 60\% masking achieved 75.7\%.
This result highlights a key difference from masked image modeling paradigms such as MAE, where masking is essential to prevent trivial reconstruction.
In contrast, our autoregressive setup naturally avoids shortcut solutions.
This suggests that random masking, while useful for preventing trivial reconstruction in pixel-space objectives, is less compatible with embedding-level prediction, where causal modeling already imposes a meaningful learning signal. 
Masking in this context introduces input corruption, disrupts sequence structure, and creates a training-inference mismatch---none of which are necessary when the prediction task is inherently non-trivial. 
These findings reinforce the appeal of our approach: prediction is sufficient, and masking, rather than helping, may obscure the signal.

\subsection{Ablation Study on Orthogonal Components}
We ablate 4 architectural components (LayerScale, RoPE, QK-Norm, and SwiGLU) introduced in Section~\ref{sec:model}. 
Starting from a plain ViT backbone, we incrementally add each component and report top-1 accuracy after 100k steps of pretraining followed by fine-tuning.
As shown in Table~\ref{tab:model_desing}, each component contributes to performance gains, and their combination yields the best result.
Notably, enabling RoPE significantly boosts accuracy, while QK-Norm further improves stability and final performance.

\vspace{-10pt}

\begin{figure*}[t]
    \centering
    \includegraphics[width=0.235\textwidth]{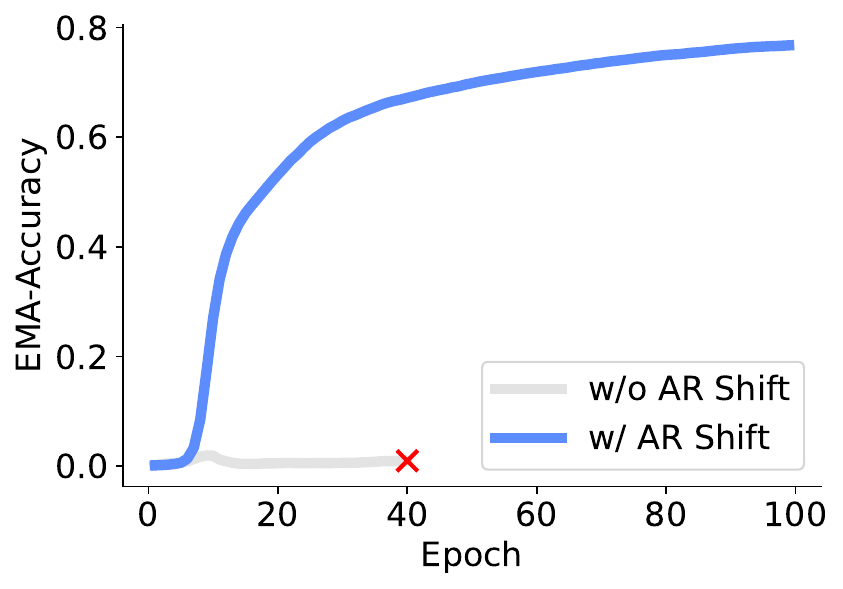}
    \includegraphics[width=0.255\textwidth]{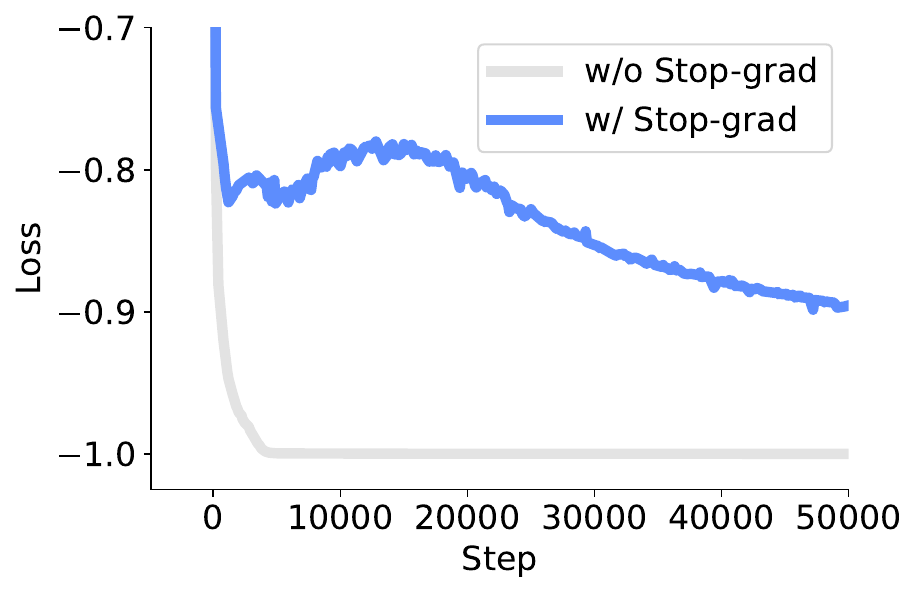}
    \includegraphics[width=0.255\textwidth]{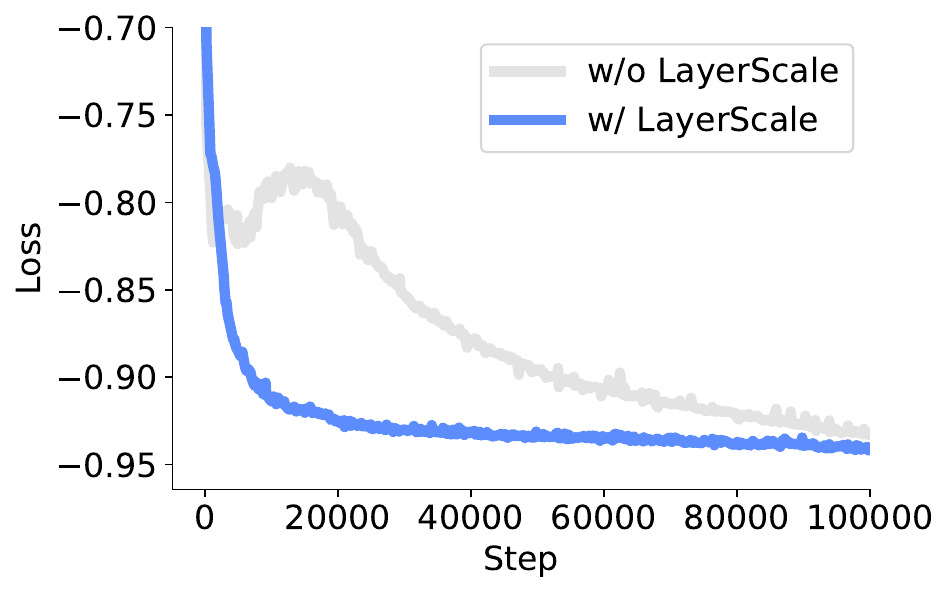}
    \includegraphics[width=0.235\textwidth]{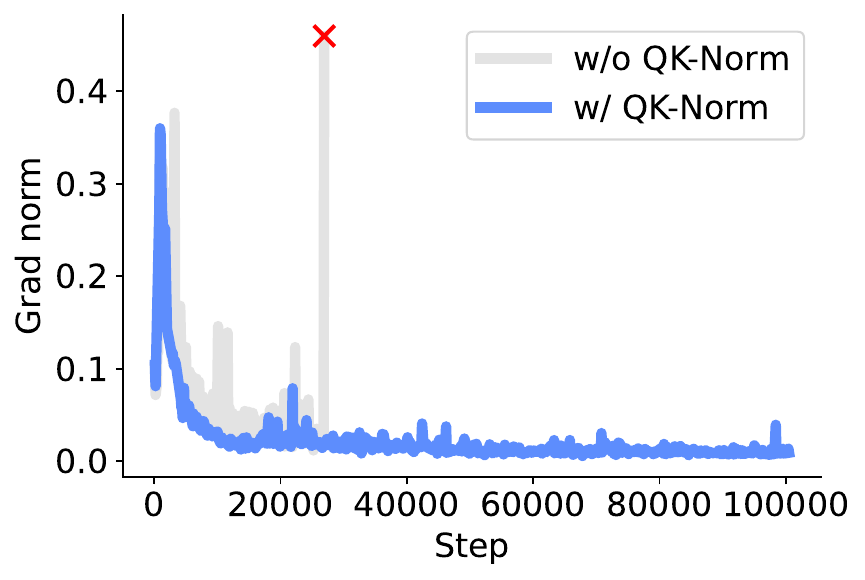}
    \vspace{-5pt}
    \caption{
        \textbf{Ablation of key components in \model pretraining.}
        \textbf{Left:} EMA accuracy with and without AR shift. Without the autoregressive shift, training diverges early.
        \textbf{Middle-left:} Training loss with and without stop-grad; removing stop-grad causes representation collapse.
        \textbf{Middle-right:} Training loss with and without LayerScale; LayerScale stabilizes optimization and accelerates convergence.
        \textbf{Right:} Gradient norm with and without QK-Norm; QK-Norm suppresses gradient explosion and improves smoothness.
    }
    \label{fig:ablation_summary}
    \vspace{-5pt}
\end{figure*}

\paragraph{LayerScale.}
We employ LayerScale to enhance training stability during pretraining. 
Although LayerScale slightly reduces top-1 accuracy on ImageNet-1K after fine-tuning, we find it plays a crucial role in stabilizing optimization and helping the model converge faster to a lower training loss during pretraining (Figure~\ref{fig:ablation_summary}). 
We attribute the degradation in fine-tuning performance to LayerScale's tendency to slow convergence in the supervised stage due to reduced gradient magnitudes.
To address this, we freeze the patch embedding layer during fine-tuning, which accelerates convergence and improves overall performance, in line with previous observations~\citep{touvron2021cait, shazeer2020glu}.
For these reasons, we retain LayerScale in our final model. 
We note that this trade-off between pretraining stability and fine-tuning speed is a common theme in modern architectures.

\vspace{-10pt}

\paragraph{RoPE.} 
We use RoPE~\citep{su2024roformer} to encode positional information in a continuous and extrapolatable manner.
Compared to absolute position embeddings, RoPE leads to a significant improvement in top-1 accuracy after fine-tuning.

\vspace{-10pt}

\paragraph{QK-Norm.} 
We use QK-Norm~\citep{henry2020qknorm} to stabilize attention and prevent gradient explosion during pretraining with SwiGLU as Section~\ref{sec:model}.
Without QK-Norm, we observe that the model becomes unstable when combined with SwiGLU, either failing to converge or requiring a significantly smaller learning rate to avoid divergence.
However, reducing the learning rate leads to underfitting and degraded performance. By applying QK-Norm to the query and key projections, we ensure smooth optimization even in deeper configurations, allowing the model to benefit from SwiGLU without sacrificing training stability.

\vspace{-10pt}

\paragraph{SwiGLU.} 
We adopt the Gated MLP variant with SwiGLU activation~\cite{shazeer2020glu}, following the design used in many recent vision~\citep{oquab2023dinov2, siméoni2025dinov3, chu2024visionllama, sun2024llamagen} and language models~\citep{touvron2023llama, bai2023qwentechnicalreport}. 
We find that replacing GeLU with SwiGLU yields only marginal performance improvements. 
Despite its limited empirical gain in our setting, we retain SwiGLU to maintain architectural alignment with modern transformer-based backbones in both vision and language domains. 
This choice ensures better compatibility with emerging model designs and facilitates future integration into unified frameworks.

\subsection{Comparisons with Previous Results}
\label{sec:comparison}

In this section, we evaluate pretrained \model checkpoints on two standard vision tasks: single-label image classification and semantic segmentation. 
For classification, we fine-tune on ImageNet-1K~\citep{russakovsky2015imagenet1k} and report top-1 accuracy. 
For segmentation, we attach a UperNet~\citep{xiao2018upernet} decoder to the \model backbone and fine-tune on ADE20K~\citep{zhou2017ade20k}, reporting mean Intersection-over-Union (mIoU).
We experiment with both \textit{Base} and \textit{Large} model variants, each pretrained on ImageNet-1K with a patch size of 14 (following DINOv2~\citep{oquab2023dinov2}) and a global batch size of 4096 (following MAE~\citep{he2022mae}). 
We use the same pretrained checkpoint for all downstream evaluations. 
Models are trained using the HuggingFace \texttt{Trainer} with distributed data parallelism (DDP) on 8 NVIDIA H100 GPUs. 
The \textit{Base} model is trained for 1600 epochs (approximately 3 days), and the \textit{Large} model for 800 epochs (approximately 5 days). 
Full training and fine-tuning details are provided in Appendix~\ref{app:training-details}.

\vspace{-10pt}

\begin{figure}[t]
    \centering
    \includegraphics[width=1.0\linewidth]{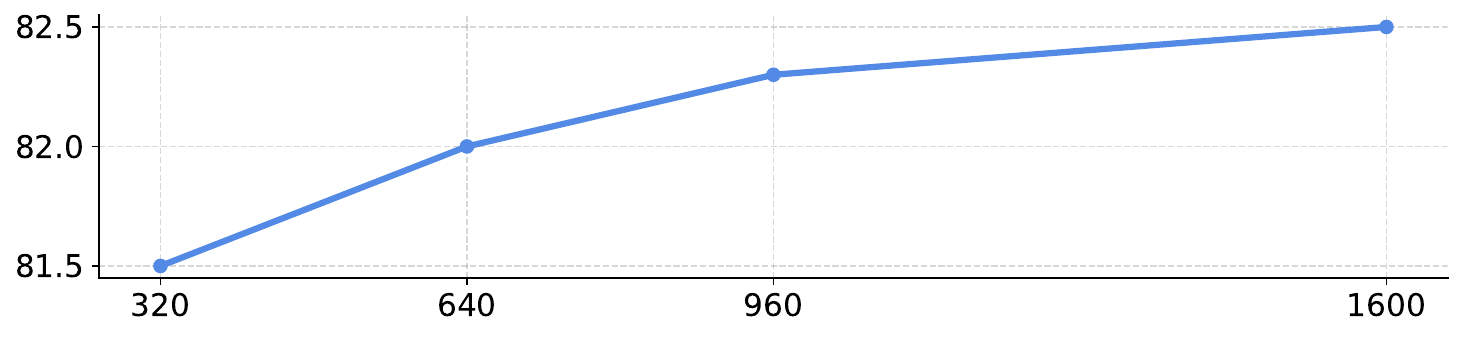}
    \includegraphics[width=1.0\linewidth]{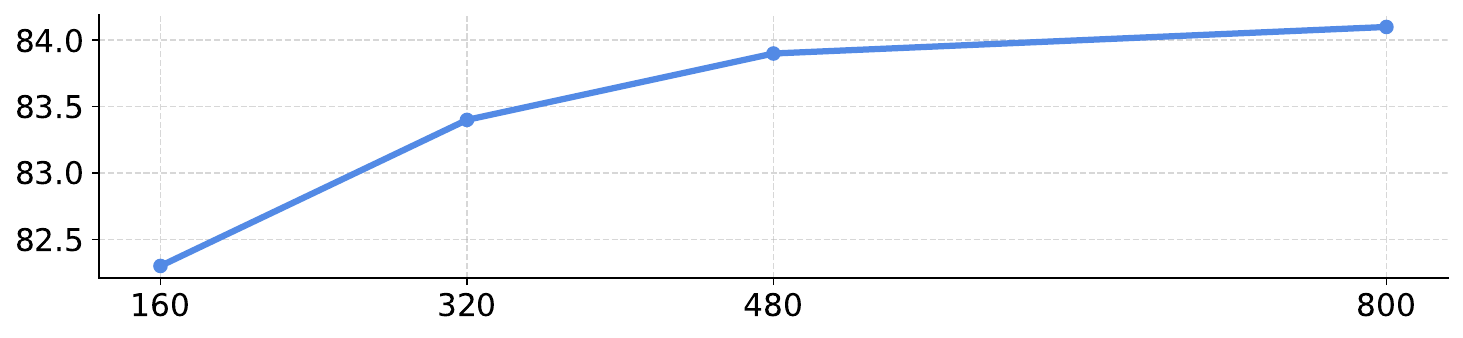}
    \caption{ImageNet-1K validation Top-1 accuracy versus training epochs. For each epoch’s checkpoint, we perform a lightweight hyperparameter search and report the best accuracy. Fine-tuning uses causal attention. The top plot corresponds to the base model, and the bottom plot to the large model.}
    \label{fig:scaling}
    \vspace{-10pt}
\end{figure}

\paragraph{Scaling behavior.}
We show that our next embedding prediction framework scales effectively with the size of the model.
As shown in Figure~\ref{fig:scaling}, model accuracy steadily improves with increased training steps, and we observe no signs of overfitting even under extended compute budgets.
Beyond verifying pretraining stability, the scaled models serve as the foundation for subsequent analysis of attention behavior and embedding structure, as well as for evaluating downstream performance via standard fine-tuning on classification and segmentation tasks.

\vspace{-10pt}

\paragraph{Fine-tuning practice.}
We find that enabling bidirectional attention during fine-tuning improves classification performance on ImageNet-1K when the model is pretrained for 100k steps, as shown in Table~\ref{tab:attn}.
Nonetheless, \model achieves competitive results even under causal attention, indicating that autoregressive embeddings retain sufficient global information for downstream tasks. 
To remain consistent with the autoregressive formulation, we use causal attention as the default setting in our experiments. 
However, we also report the results of bidirectional attention for completeness. 
For semantic segmentation, we adopt bidirectional attention during fine-tuning by default, as each output embedding corresponds to a localized prediction and requires access to the full spatial context of the input image. 

\vspace{-10pt}

\paragraph{Classification results.}
We follow standard protocols~\citep{touvron2021deit, bao2022beit, he2022mae} and fine-tune \model on the labeled ImageNet-1K dataset. 
We use similar hyperparameters as MAE~\citep{he2022mae}, including layer-wise learning rate decay~\citep{clark2020llrd, bao2022beit}, RandAugment~\citep{dubuk2020randaugement}, label smoothing~\citep{szegedy2016label}, mixup~\citep{zhang2018mixup}, cutmix~\citep{yun2019cutmix}, and DropPath~\citep{huang2016droppath}. 
Results are summarized in Table~\ref{tab:imagenet1k}.
Compared to prior pretraining methods such as MAE~\citep{he2022mae}, BEiT~\citep{bao2022beit}, and MoCo~\citep{he2020moco}, \model achieves competitive top-1 accuracy, reaching 83.8\% on the \textit{Base} model and 85.3\% on the \textit{Large} model.
Our method does not require any task-specific head, decoder, or auxiliary loss. 
Unlike contrastive or masked prediction approaches, we adopt a single-stream autoregressive formulation with one forward pass, no reconstruction target, and no multi-branch architecture. 
Moreover, the strong performance under causal attention demonstrates the effectiveness of embedding-level autoregression in capturing transferable semantics.

\begin{table*}[ht]
\centering
\scriptsize
\renewcommand{\arraystretch}{1.2}
\resizebox{0.98\textwidth}{!}{
\begin{tabular}{l l l l c c c}
\hline
\textbf{Model} & \textbf{Pretrain Task} & \textbf{Pretrain Framework} & \textbf{Decoder} & \textbf{\# FWD / step} & \textbf{Epochs} & \textbf{Acc~(\%)} \\
\hline
\multicolumn{7}{l}{\textbf{ViT-B}} \\
\hline
MoCo v3-B~\citep{chen2021mocov3}  & contrastive learning      & siamese              & mlp proj.\ head          & 2 & 600  & 83.2 \\
BEiT-B~\citep{bao2022beit}        & masked token pred                & masked modeling   & linear pred.\ head                  & 1 & 800  & 83.4 \\
DINO-B~\citep{caron2021dino}      & self-distillation      & siamese              & mlp proj.\ head                     & N & 1600 & \underline{83.6} \\
MAE-B~\citep{he2022mae}           & masked pixel pred                & masked autoencoder   & transformer decoder                & 1 & 1600 & \underline{83.6} \\
\rowcolor{cyan!15}
\textbf{\model-B}$^*$             & autoreg.\ embed pred      & autoregression       & none                                 & 1 & 1600 & 82.5 \\
\rowcolor{cyan!15}
\textbf{\model-B}                 & autoreg.\ embed pred      & autoregression       & none                                 & 1 & 1600 & \textbf{83.8} \\
\hline
\multicolumn{7}{l}{\textbf{ViT-L}} \\
\hline
MoCo v3-L~\citep{chen2021mocov3}  & contrastive learning      & siamese              & mlp proj.\ head          & 2 & 600  & 84.1 \\
iBot-L~\citep{zhou2022ibot}  & self-dist \& masked token pred    & siamese \& masked modeling             & mlp proj.\ head          & 4 & 1000  & 84.8 \\
BEiT-L~\citep{bao2022beit}        & masked token pred                & masked modeling   & linear pred.\ head                  & 1 & 800  & 85.2 \\
MAE-L~\citep{he2022mae}           & masked pixel pred                & masked autoencoder   & transformer decoder                & 1 & 1600 & \textbf{85.6}$^{\dagger}$ \\
JEPA-L~\citep{assran2023jepa}     & masked embed pred                & siamese \& masked modeling & transformer predictor            & 2 & 300  & 85.2$^{\dagger}$ \\
\rowcolor{cyan!15}
\textbf{\model-L}$^*$             & autoreg.\ embed pred      & autoregression       & none                                 & 1 & 800  & 84.1 \\
\rowcolor{cyan!15}
\textbf{\model-L}                 & autoreg.\ embed pred      & autoregression       & none                                 & 1 & 800  & \underline{85.3} \\
\hline
\end{tabular}
}
\vspace{-5pt}
\caption{Comparison of different self-supervised learning frameworks on ImageNet-1K classification. Results are grouped by model scale, with Base models in the upper block and Large models in the lower block. Effective pretraining epochs are used based on the actual number of images or views seen during training; see~\citep{zhou2022ibot} for details. $^*$ indicates methods that use causal attention during fine-tuning. $^{\dagger}$ Denotes results based on our implementation.
}
\label{tab:imagenet1k}
\vspace{-10pt}
\end{table*}
\begin{table}[ht]
\centering
\scriptsize
\renewcommand{\arraystretch}{1.2}
\resizebox{0.9\linewidth}{!}{
\begin{tabular}{l l c c}
\hline
\textbf{Method} & \textbf{Pre-train data} & \textbf{ViT-B} & \textbf{ViT-L} \\
\hline
Supervised & IN1K w/ labels & 47.4 & 49.9 \\
MoCo v3~\citep{chen2021mocov3}    & IN1K           & 47.3 & 49.1 \\
BEiT~\citep{bao2022beit}       & IN1K + DALLE   & 47.1 & 53.3 \\
MAE~\citep{he2022mae}        & IN1K           & \underline{48.1} & \underline{53.6} \\
\rowcolor{cyan!15}
\textbf{\model} & IN1K         & \textbf{48.3} & \textbf{54.0} \\
\hline
\end{tabular}
}
\vspace{-5pt}
\caption{Comparison of ADE20K semantic segmentation (mIoU) under different pretraining methods.}
\label{tab:ade20k}
\vspace{-10pt}
\end{table}

\vspace{-10pt}

\paragraph{Semantic Segmentation.}
We evaluate \model on ADE20K by attaching a standard UperNet~\citep{xiao2018upernet} decoder and fine-tuning the model using bidirectional attention, as described before. 
ADE20K is a challenging scene parsing benchmark with dense pixel-wise annotations. 
We follow the training recipe from MMSegmentation~\citep{mmseg2020}, using a crop size of $512 \times 512$, batch size of 16, and 160K training iterations.
As shown in Table~\ref{tab:ade20k}, \model achieves competitive performance with only ImageNet-1k pretraining, reaching 48.3\% mIoU for \textit{Base} model and 54.0\% mIoU for \textit{Large} model.
Despite not using any decoder or pixel-level objective during pretraining, \model transfers effectively to dense prediction tasks. 
Unlike prior work relying on contrastive learning or pixel reconstruction, our method learns transferable representations purely through next embedding prediction.

\section{Quantitative Results}

In this section, we investigate how \model organizes visual information by analyzing its attention patterns and the structure of its learned embeddings. 
Our goal is to understand whether the next-embedding prediction objective induces meaningful global dependencies and semantic organization.

\subsection{Attention Map Analysis}

To better understand how \model exploits context when predicting future patches, we visualize representative attention maps on ImageNet-1K images in Figure~\ref{fig:attn_im1k}.
For each triplet, the first column marks the current query patch in the original image, and the second column shows the corresponding attention maps from \model.

Interestingly, the attention maps are often long-ranged and object-centric. 
Across a wide range of categories and viewpoints, the model consistently allocates most of its attention to regions that are semantically related to the query patch. 
These regions typically include other parts of the same object or nearby informative structures, rather than uniformly attending to all patches or focusing only on local neighbors. 
For example, when the query patch lies on the head of an animal or the body of a person, the model tends to attend to other body parts and task-relevant background regions, even when they are spatially distant.
In cluttered scenes, \model suppresses distractors and concentrates on a small subset of visually coherent patches.
These qualitative results suggest that the predictive objective encourages the Transformer to form global, semantically meaningful dependencies between patches, effectively learning to localize and group object parts without any explicit supervision.

\subsection{Embedding Analysis}

\begin{figure*}[t]
    \centering
    \begin{subfigure}{\linewidth}
        \centering
        \includegraphics[width=\linewidth]{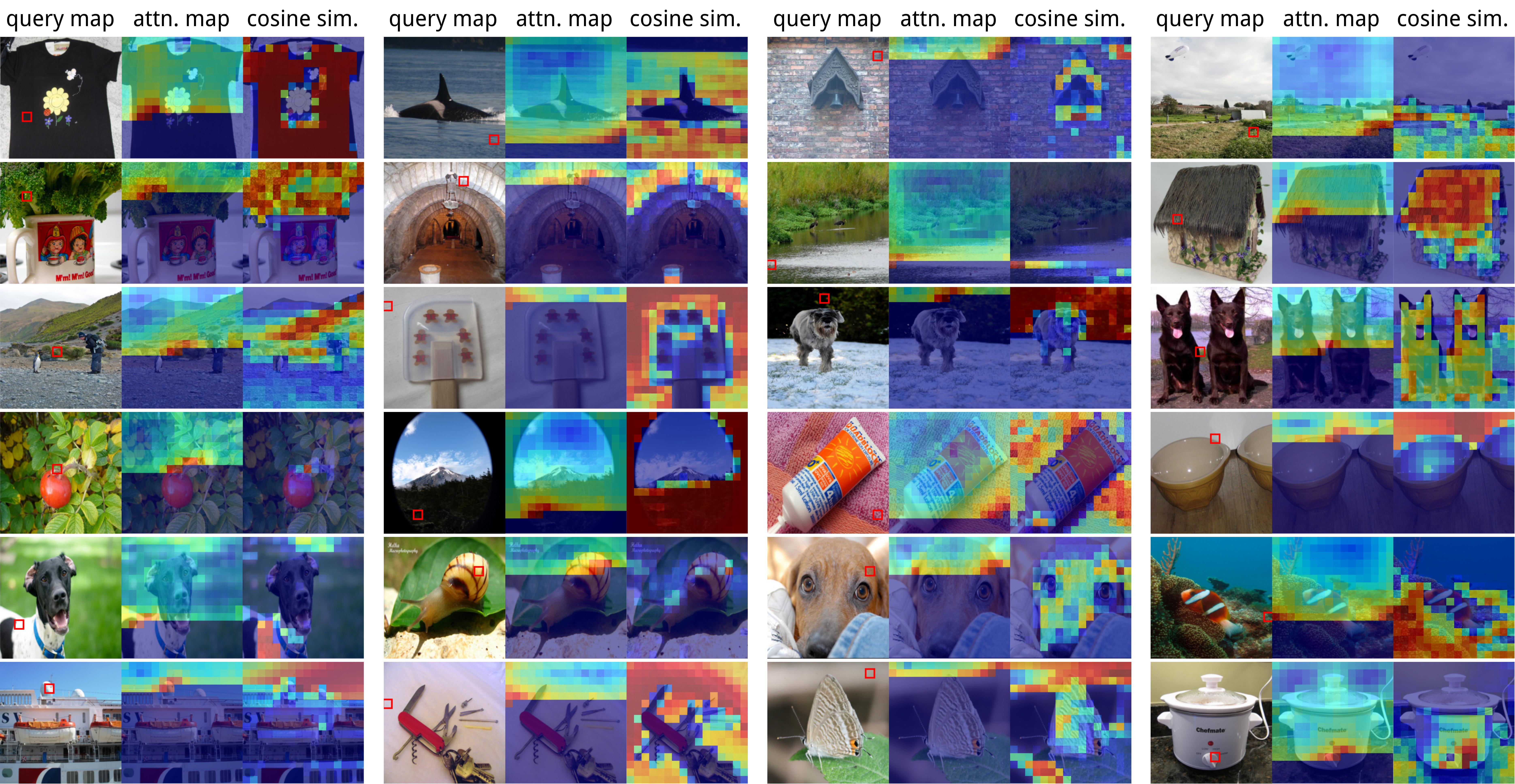}
        \caption{ImageNet-1K validation samples (unseen during pretraining).}
        \label{fig:attn_im1k}
    \end{subfigure}
    ~
    \begin{subfigure}{\linewidth}
        \centering
        \includegraphics[width=\linewidth]{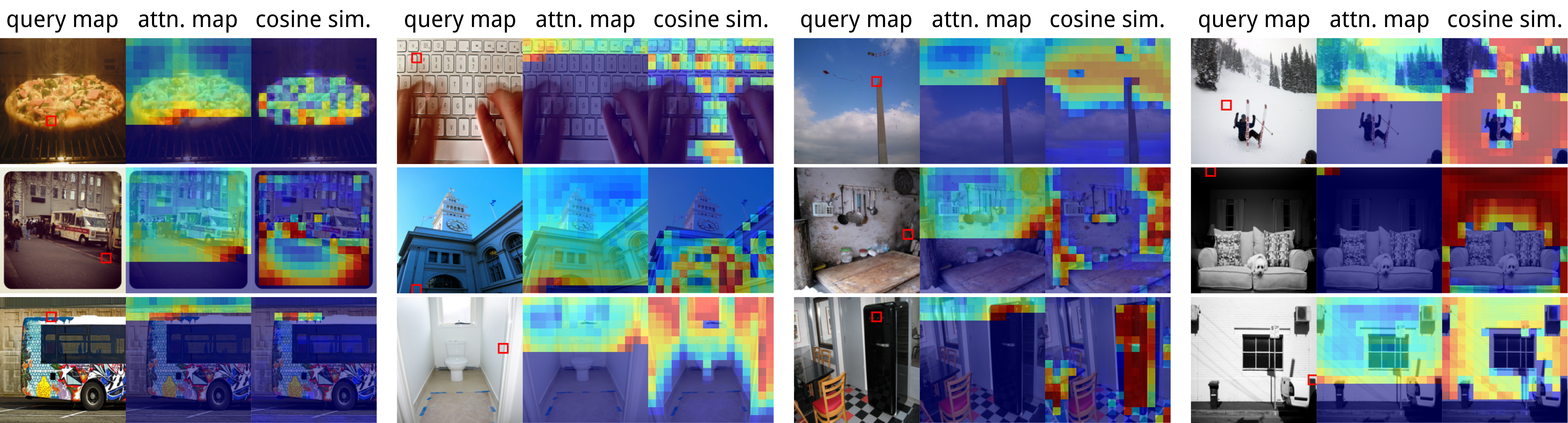}
        \caption{MSCOCO validation samples (out of distribution during pretraining).}
        \label{fig:attn_unseen}
    \end{subfigure}
    \vspace{-15pt}
    \caption{
    \textbf{Attention and embedding analyses.}
    Each example consists of three views:  
    (i) the \textbf{query patch} (\textcolor{red}{\textbf{$\square$}}) highlighted in the original image,  
    (ii) the \textbf{attention map} from the \model showing which patches the model attends to when predicting the next embedding, and  
    (iii) the \textbf{embedding-similarity map} showing the cosine similarity between the predicted embedding and all other patch embeddings in the same image.  
    Warmer colors indicate higher attention or greater similarity; cooler colors indicate lower values.
    }
    \vspace{-12pt}
\end{figure*}

We further analyze the embeddings produced by \model by examining the similarity between the predicted embedding of the next patch and all other patch embeddings within the same image, shown in the third column of each triplet in Figure~\ref{fig:attn_im1k}. 
We observe that the predicted embedding is most similar to patches that belong to the same object or semantic region as the current patch, while unrelated background areas exhibit much lower similarity.
This behavior emerges even though \model is never trained with explicit labels or region annotations.

The embedding similarity maps reveal that the model does not merely memorize local texture; instead, it learns to extrapolate object-level structure.
When the query patch lies on a distinctive part of an object, the predicted embedding is highly similar to distant patches covering other parts of that object, and often to patches that are occluded or outside the local neighborhood.
In homogeneous backgrounds, the similarity patterns become more diffuse but remain coherent, reflecting uncertainty in the visual context.
Overall, these observations indicate that generative pretraining with next-embedding prediction encourages \model to learn embeddings that are both predictive and semantically organized, which likely contributes to its strong downstream transfer performance.

\vspace{-5pt}

\section{Conclusion and Future Work}

This work revisits the core idea of causal next-token prediction, not in pixel or token space, but in the embedding space of vision models. We demonstrate that simple next embedding prediction is sufficient to learn transferable visual representations that scale.
By treating patch embeddings as prediction targets, we avoid brittle design choices tied to handcrafted pretext tasks, and instead rely on the structure implicitly learned through the sequence. With only IN1K self-supervised pretraining, \model matches or slightly outperforms mainstream methods on classification and segmentation transfer, while using a simpler training paradigm (single forward pass, no decoder). What emerges is not just an algorithm, but a different perspective: that the simplicity of autoregression, when properly adapted to vision, can help unify pretraining paradigms across \mbox{modalities.}

\vspace{-10pt}

\paragraph{Modality-agnostic potentials.}
Recent Large Language Models increasingly adopt tied embeddings, where the input and output embedding matrices are shared. 
This mechanism is effectively equivalent to predicting the next embedding in the latent space, which is precisely the principle underlying our framework. 
From this perspective, our work does not introduce a new paradigm, but rather reveals a unifying view: different modalities can be trained under the same objective, with embeddings serving as a common representational currency.

\vspace{-10pt}

\paragraph{Generative potentials.}
At the same time, our formulation naturally extends toward generative modeling. By coupling \model with a suitable image decoder or diffusion-based generator~\citep{goodfellow2020gan,ho2020ddpm,song2021ddim,peebles2023dit}, the same autoregressive embedding predictor could be used for image synthesis or editing. Exploring this direction, which bridges representation learning and generation within a unified model, remains an exciting avenue for future work.

\vspace{-2pt}

\subsection*{Acknowledgment}
This work is supported in part by a departmental fellowship to Sihan Xu and Weinberg Cognitive Science Fellowship to Sihan Xu and Ziqiao Ma.
We acknowledge the computing resources of Google Cloud Platform and Lambda Labs.
We sincerely thank Steven Bucaille for opening a pull request that adds NEPA to the \texttt{transformers} ecosystem.
We thank Jiayi Pan and Freda Shi for their valuable feedback.
\clearpage
\setcounter{page}{1}
\maketitlesupplementary
\appendix

\section{Broader Impacts}
\model is a generic self-supervised visual pretraining method and, like other large models, may inherit dataset biases or be misused to create misleading visual content, so care is needed in data selection, and deployment, including appropriate safeguards and monitoring.

\section{Methodology Comparisons}

\paragraph{Relation to CPC~\citep{oord2018cpc}.}
CPC-style methods mainly train an encoder with a contrastive loss, where a small autoregressive module just aggregates context. 
In \model, the main learned object is the transformer predictor itself: we use a simple shared embedding layer and directly regress the next embedding without negatives or a contrastive head, which makes the architecture easier to scale.

\vspace{-10pt}

\paragraph{Relation to GPT~\cite{radford2018gpt}, iGPT~\citep{chen2020igpt} and AIM~\cite{el2024scalable,fini2025multimodal}.}
\model shares the causal prediction idea, but works fully in the continuous embedding space. 
We do not use task-specific decoders, tokenizers, or language heads during pretraining.
iGPT and AIM-style methods supervise against predefined pixel-space targets (discrete pixel tokens or continuous patch values), whereas \model supervises in representation space via embedding similarity, a learned, moving target defined by the embedding rather than pixels.

\vspace{-10pt}

\begin{figure}[h]
    \centering
    \includegraphics[width=0.9\linewidth]{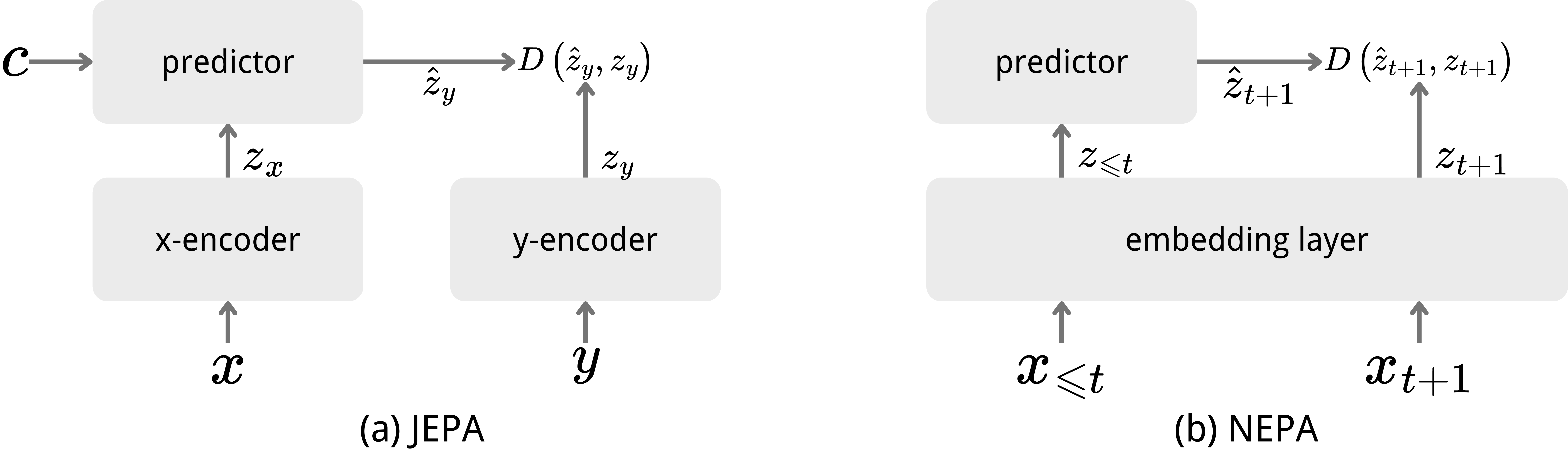}
    \caption{\textbf{Comparison between JEPA and NEPA.}}
    \vspace{-10pt}
    \label{fig:jepa_nepa}
    \vspace{-5pt}
\end{figure}

\paragraph{Relation to JEPA~\citep{assran2023jepa}.}
JEPA uses separate encoders for context and target views plus a heavy head to score representation pairs. As shown in Fig.~\ref{fig:jepa_nepa}, \model keeps the JEPA-style latent prediction goal but simplifies the architecture to a single embedding layer and an autoregressive transformer predictor, without asymmetric branches or an extra head.

\section{Implementation Details}
\label{app:impl-details}

We first prototyped and validated \model using the \texttt{timm}~\citep{rw2019timm} library. 
For large-scale experiments, we switched to the Hugging Face ecosystem, relying on \texttt{transformers}~\citep{wolf-etal-2020-transformers}, \texttt{datasets}~\citep{lhoest-etal-2021-datasets}, and \texttt{evaluate}. 
Our implementation is mainly based on the ViT code in \texttt{transformers.models.vit.modeling\_vit}, and the training scripts are adapted from the official \texttt{run\_mim} and \texttt{run\_image\_classification} examples. 
For data augmentation, we use the \texttt{timm} utilities, in particular \texttt{Mixup} and \texttt{create\_transform}. All ImageNet experiments use the \texttt{ILSVRC/imagenet-1k} dataset accessed through Hugging Face \texttt{datasets}. 
ADE20K experiments follow the standard recipes and data processing pipelines from \texttt{mmsegmentation}~\citep{mmseg2020}.\footnote{Our code and model is available at \url{https://github.com/SihanXU/nepa}.}

\section{Training Details}
\label{app:training-details}

\paragraph{Pre-training.}
The full pre-training hyperparameters are listed in Table~\ref{tab:pretrain_setting}. 
For all models we use the standard linear learning-rate scaling rule, i.e., $\text{lr} = \text{base\_lr} \times B / 256$,
where $B$ is the global batch size.
\vspace{-10pt}
\begin{table}[h]
\centering
    \begin{tabular}{l l}
    config & value \\
    \hline
    optimizer & AdamW \cite{loshchilov2018adamw} \\
    base learning rate & 3e-4 \\
    weight decay & 0.05 \\
    optimizer momentum & $\beta_1, \beta_2{=}0.9, 0.95$ \cite{chen2020igpt} \\
    batch size & 4096 \\
    learning rate schedule & cosine decay \\
    warmup epochs \cite{goyal2018warmup} & 40 \\
    augmentation & RandomResizedCrop \\
    \end{tabular}
    \vspace{-5pt}
    \caption{\textbf{Pre-training setting.}}
    \label{tab:pretrain_setting}
\vspace{-10pt}
\end{table}

\paragraph{End-to-end Fine-tuning.}
Fine-tuning settings are summarized in Table~\ref{tab:finetune_setting}. We use the same linear learning-rate scaling rule as in pre-training and apply layer-wise learning-rate decay. Instead of using a fixed decay factor, we linearly increase the decay rate from a smaller value to $1.0$ over the course of training, which helps reduce overfitting at the beginning while still allowing all layers to adapt later.

\begin{table}[h]
\centering
\small
    \begin{tabular}{l l}
    config & value \\
    \hline
    optimizer & AdamW \\
    base learning rate & 1e-3 \\
    weight decay & 0.05 \\
    optimizer momentum & $\beta_1, \beta_2{=}0.9, 0.999$ \\
    layer-wise lr decay \cite{bao2022beit, he2022mae} & $0.35$ (B) $0.60$ (L) \\
    batch size & 1024 \\
    learning rate schedule & cosine decay \\
    warmup epochs & 5 \\
    training epochs & 100 (B), 50 (L) \\
    augmentation & RandAug (9, 0.5) \cite{touvron2021deit,bao2022beit,he2022mae} \\
    label smoothing \cite{szegedy2016label} & 0.1 \\
    mixup \cite{zhang2018mixup} & 0.8 \\
    cutmix \cite{yun2019cutmix} & 1.0 \\
    drop path \cite{huang2016droppath} & 0.1 (B) 0.2 (L) \\
    \end{tabular}
    \vspace{-5pt}
    \caption{\textbf{Fine-tuning setting.}}
    \label{tab:finetune_setting}
\vspace{-10pt}
\end{table}

\section{Additional Experiments}
\label{app:add-exps}

\subsection{Training Dynamics}

\begin{figure}[h]
    \centering
    \includegraphics[width=1.0\linewidth]{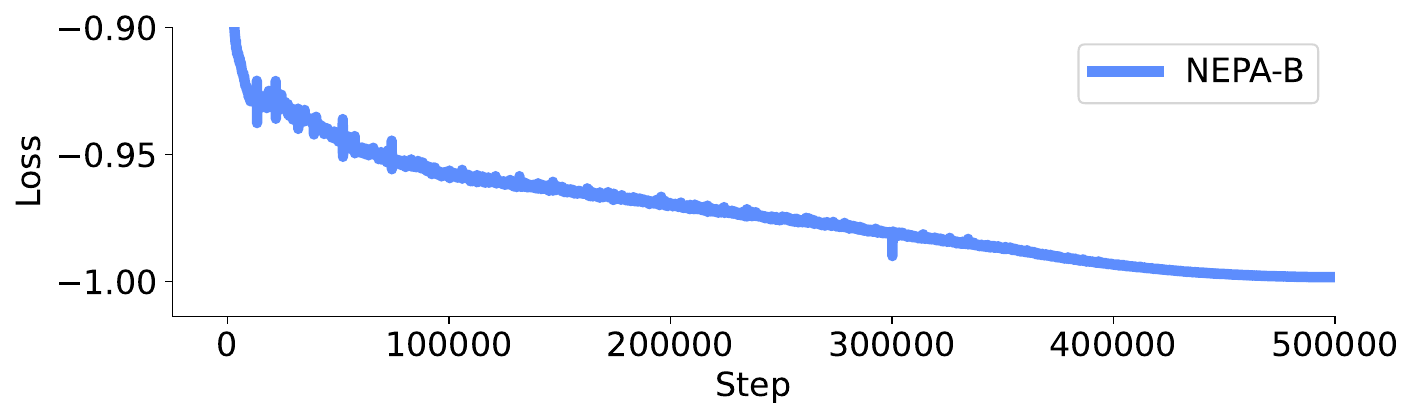}
    \vspace{-2mm}
    \includegraphics[width=1.0\linewidth]{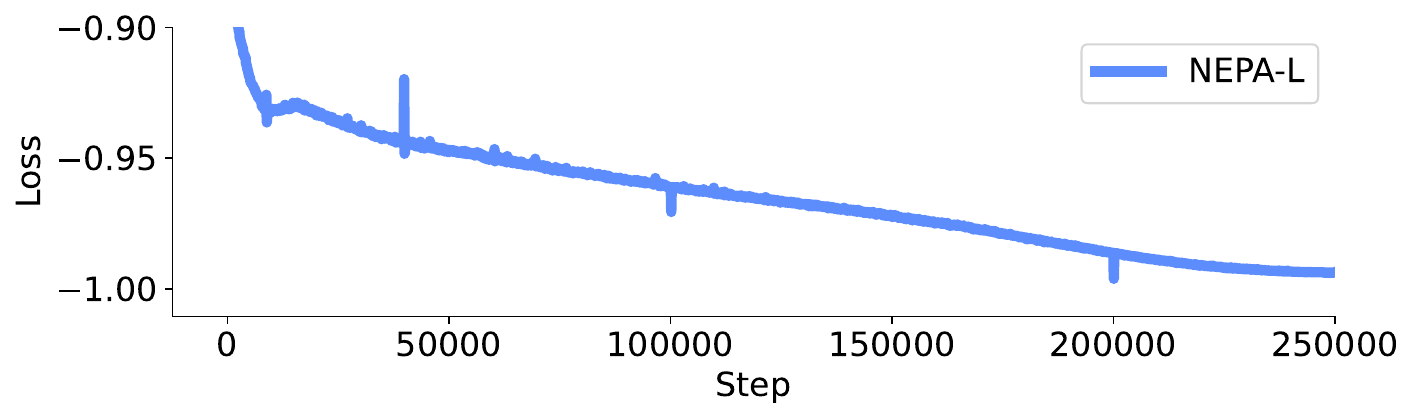}
    \vspace{-2mm}
    \includegraphics[width=1.0\linewidth]{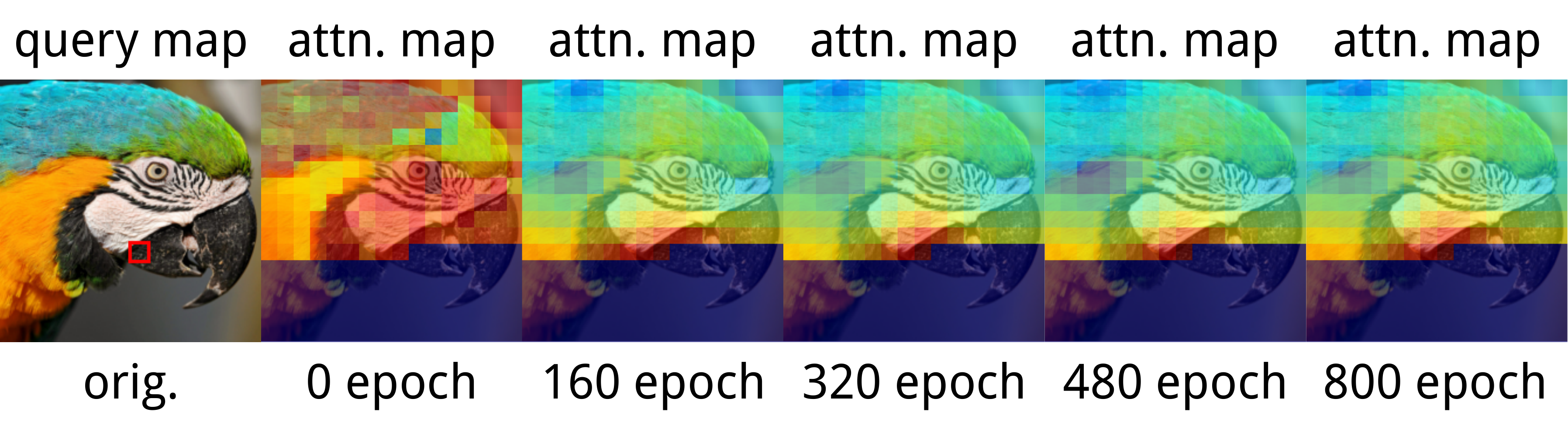}
    \vspace{-2mm}
    \caption{\textbf{Pre-training dynamics of NEPA.} Top: pre-training loss curves;
    Bottom: evolution of NEPA-L attention maps.}
    \label{fig:loss_curve}
    \vspace{-10pt}
\end{figure}

We report the pre-training loss curves and the attention maps from \model-L of intermediate checkpoints in this section (Figure~\ref{fig:loss_curve}). The plots and figures illustrate the overall training stability of \model and how the attention evolves as pre-training progresses.

\subsection{Limitations and Failure Cases}

\paragraph{Linear Probing.}

We evaluated two variants for linear probing, one using the last embedding of the autoregressive model and one using the average of all embeddings, with results reported in Table~\ref{tab:probing}. 
\model performs poorly under standard linear probing. 
This is expected, as the output representation is very close to the features right after the embedding layer, thus the probing result mainly reflects this shallow representation rather than the full capacity of the predictor.

\begin{table}[h]
\centering
\footnotesize
\begin{tabular}{c c c}
\toprule
\textbf{\model-B} & Last Embed & Avg. Embed \\
\midrule
{\textbf{Acc}~(\%)} & 11.3 & 14.1\\
\bottomrule
\end{tabular}
\vspace{-2mm}
\caption{\textbf{Linear probing on \model-B.} Top-1 accuracy using the last autoregressive embedding or the average over all embeddings.}
\label{tab:probing}
\vspace{-10pt}
\end{table}

\vspace{-10pt}

\paragraph{Quantitative Failure Examples.}

\begin{figure}[t]
    \centering
    \includegraphics[width=1.0\linewidth]{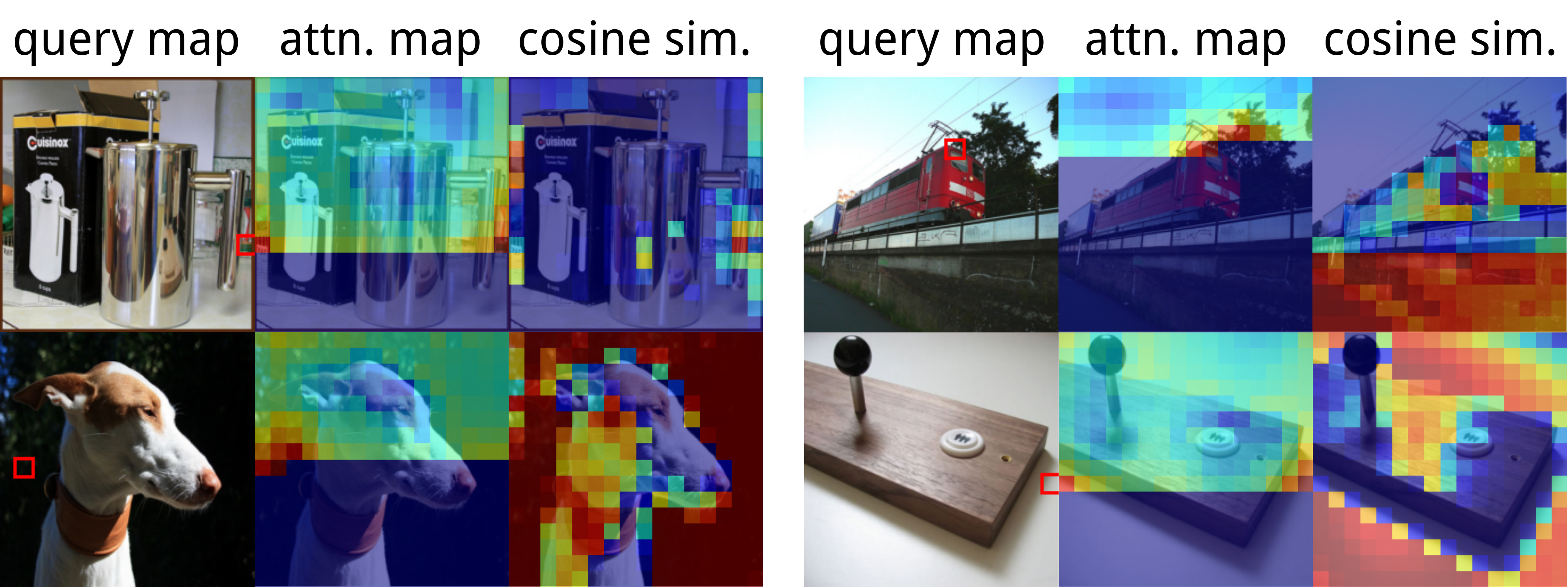}
    \caption{\textbf{Quantitative Failure Examples.}
    In a multi-object scene with strong reflections, the model confuses highlights on metal surfaces with object embeddings; under backlighting, shadowed regions are misinterpreted as trees; for the animal, shaded skin is treated as background; and in the last example, bright reflective regions of the object are also mistaken for background.}
    \label{fig:failure}
    \vspace{-10pt}
\end{figure}

We also inspect typical failure cases in Figure~\ref{fig:failure}. 
Under the current scale, \model often struggles with images that require non-trivial and reasoning-intensive physical understanding, such as interpreting reflections, shading, and shadows, as well as scenes containing many small or overlapping objects. 
In such cases, the model tends to produce uncertain or inconsistent predictions, suggesting room for improvement in reasoning about complex spatial layouts.
We hypothesize that this reflects the limitation of the ImageNet dataset on which we train our current model~\cite{taesiri2023imagenet,neuhaus2023spurious}, and this could be addressed as we scale up with more diverse datasets.

\subsection{Ablations on MAE Baseline Components}
Table~\ref{tab:ablation_mae} reports the effects of adding our orthogonal components to a MAE baseline, where they bring little or no gain in our experiments. 
We hypothesize masked reconstruction is less sensitive to embedding/positional issues, RoPE mainly helps autoregressive models, and QK-Norm is only useful when paired with LayerScale.

\begin{table}[h]
\centering
\scriptsize
\begin{tabular}{c c c c c}
\toprule
\textbf{MAE} & Our impl & w/ LayerScale & w/ RoPE & w/ QK-Norm \\
\midrule
{\textbf{Acc}~(\%)} & 85.6 & 85.5 & 85.6 & 85.6\\
\bottomrule
\end{tabular}
\caption{\textbf{Ablations on MAE with our components.} Applying LayerScale, RoPE, and QK-Norm to a strong MAE baseline brings little additional gain on performance.}
\label{tab:ablation_mae}
\vspace{-10pt}
\end{table}

\subsection{Layer-wise LR decay and Overfitting}
We study the effect of our layer-wise learning-rate decay (LLRD) schedule on overfitting. Instead of using a fixed decay factor, we linearly increase the decay from a small initial value to $1.0$ over the course of fine-tuning, effectively unfreezing deeper layers gradually. As shown in Table~\ref{tab:overfit_llrd}, this schedule reduces overfitting and improves performance: on the \model-B backbone the accuracy increases from $83.0\%$ to $83.8\%$.
\begin{table}[h]
\centering
\begin{tabular}{c c c}
\toprule
\textbf{Layer-wise LR decay} & $0.65$ & $0.35 \rightarrow 1.00$ \\
\midrule
{\textbf{Acc}~(\%)} & 83.0 & 83.8 \\
\bottomrule
\end{tabular}
\caption{\textbf{Effect of layer-wise LR decay on \model-B.} Comparing a fixed decay factor of $0.65$ with a schedule that increases from $0.35$ to $1.00$ during fine-tuning.}
\label{tab:overfit_llrd}
\vspace{-15pt}
\end{table}

\begin{figure*}[h]
    \centering
    \includegraphics[width=\linewidth]{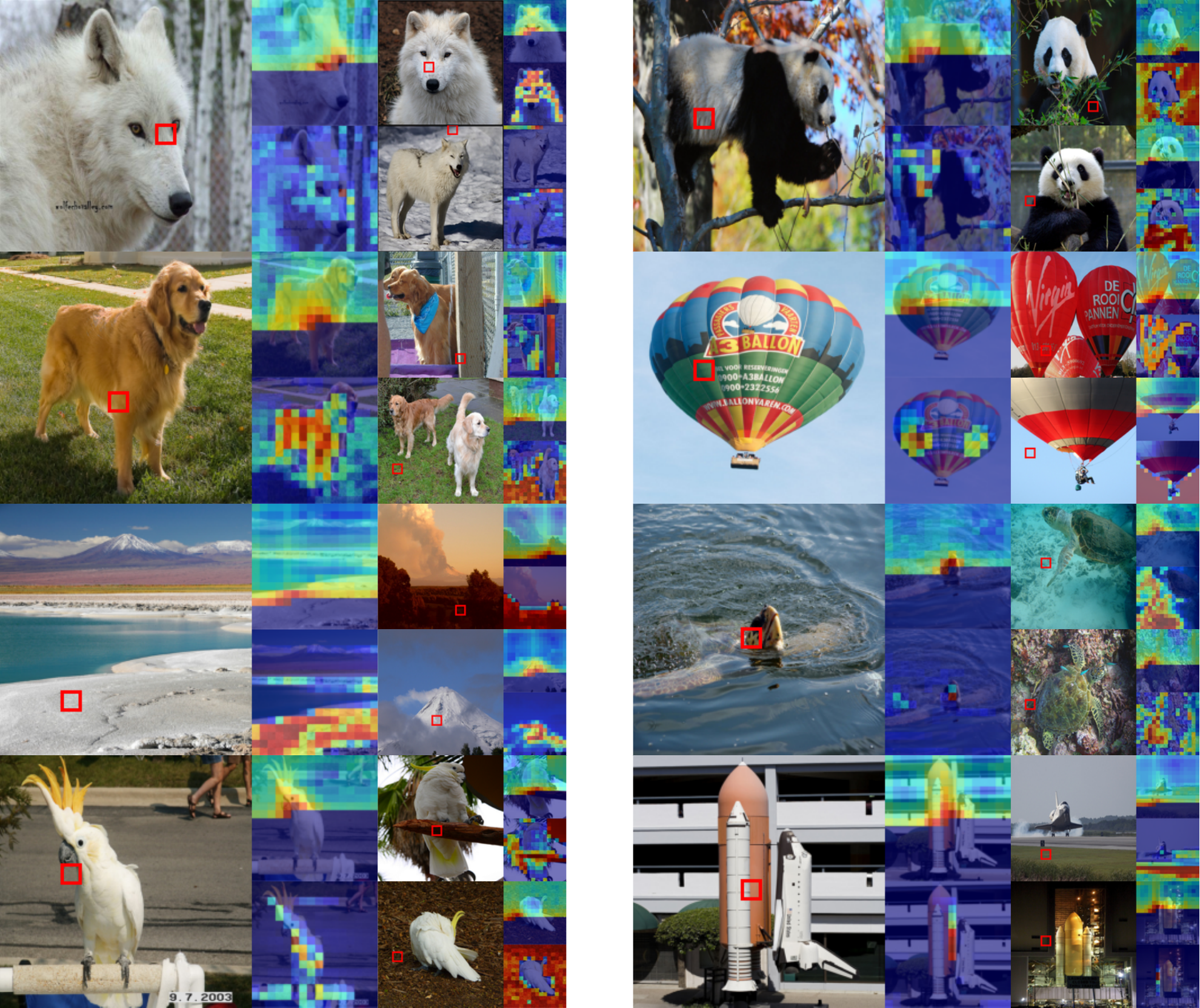}
    \caption{\textbf{Additional attention and embedding visualizations.}}
    \label{fig:additional_vis}
    \vspace{-10pt}
\end{figure*}

\subsection{Reproduced Results without SwiGLU}
Finally, we evaluate \model without SwiGLU, replacing it with standard feed-forward layers. For this comparison we lightly tune the learning rate and layer-wise LR decay and report the best setting. Table~\ref{tab:reproduce} shows that we can reproduce similar results in this configuration, indicating that SwiGLU is non-essential optimization and that the main gains come from the overall predictive architecture.
\begin{table}[h]
\centering
\begin{tabular}{c c c}
\toprule
\textbf{\model-B} & w/ SwiGLU & w/ GeLU \\
\midrule
{\textbf{Acc}~(\%)} & 83.8 & 83.6 \\
\bottomrule
\end{tabular}
\caption{\textbf{Reproduced Results of \model-B without SwiGLU.}}
\label{tab:reproduce}
\vspace{-10pt}
\end{table}

\subsection{Additional Quantitative Results}

Figure~\ref{fig:additional_vis} shows additional attention and embedding visualizations for eight ImageNet classes: white wolf (270), giant panda (388), golden retriever (207), balloon (417), volcano (980), loggerhead turtle (33), sulphur-crested cockatoo (89), and space shuttle (812). For each query patch, we display the query map, the attention map over all seen patches, and the cosine-similarity map between the predicted and true embeddings. Across these diverse classes, the model consistently attends to semantically related regions and assigns high similarity to patches on the same object, supporting our claim that \model learns object-centric and spatially coherent representations.

{
    \small
    \bibliographystyle{ieeenat_fullname}
    \bibliography{main}

@String(IJCV = {Int. J. Comput. Vis.})

@String(CVPR= {IEEE Conf. Comput. Vis. Pattern Recog.})

@String(ICCV= {Int. Conf. Comput. Vis.})

@String(ECCV= {Eur. Conf. Comput. Vis.})

@String(IJCV  = {IJCV})

@String(CVPR  = {CVPR})

@String(ICCV  = {ICCV})

@String(ECCV  = {ECCV})

@inproceedings{devlin019bert,
    title = "{BERT}: Pre-training of Deep Bidirectional Transformers for Language Understanding",
    author = "Devlin, Jacob  and
      Chang, Ming-Wei  and
      Lee, Kenton  and
      Toutanova, Kristina",
    editor = "Burstein, Jill  and
      Doran, Christy  and
      Solorio, Thamar",
    booktitle = "Proceedings of the 2019 Conference of the North {A}merican Chapter of the Association for Computational Linguistics: Human Language Technologies, Volume 1 (Long and Short Papers)",
    month = jun,
    year = "2019",
    address = "Minneapolis, Minnesota",
    publisher = "Association for Computational Linguistics",
    url = "https://aclanthology.org/N19-1423/",
    doi = "10.18653/v1/N19-1423",
    pages = "4171--4186",
}

@INPROCEEDINGS{wang2015video,
author={Wang, Xiaolong and Gupta, Abhinav},
booktitle={2015 IEEE International Conference on Computer Vision (ICCV)}, 
title={Unsupervised Learning of Visual Representations Using Videos}, 
year={2015},
volume={},
number={},
pages={2794-2802},
doi={10.1109/ICCV.2015.320}}

@InProceedings{noroozi2016puzzles,
author="Noroozi, Mehdi
and Favaro, Paolo",
editor="Leibe, Bastian
and Matas, Jiri
and Sebe, Nicu
and Welling, Max",
title="Unsupervised Learning of Visual Representations by Solving Jigsaw Puzzles",
booktitle="Computer Vision -- ECCV 2016",
year="2016",
publisher="Springer International Publishing",
address="Cham",
pages="69--84",
isbn="978-3-319-46466-4"
}

@InProceedings{zhang2016color,
author="Zhang, Richard
and Isola, Phillip
and Efros, Alexei A.",
editor="Leibe, Bastian
and Matas, Jiri
and Sebe, Nicu
and Welling, Max",
title="Colorful Image Colorization",
booktitle="Computer Vision -- ECCV 2016",
year="2016",
publisher="Springer International Publishing",
address="Cham",
pages="649--666",
isbn="978-3-319-46487-9"
}

@InProceedings{pathak2017move,
author = {Pathak, Deepak and Girshick, Ross and Dollar, Piotr and Darrell, Trevor and Hariharan, Bharath},
title = {Learning Features by Watching Objects Move},
booktitle = {Proceedings of the IEEE Conference on Computer Vision and Pattern Recognition (CVPR)},
month = {July},
year = {2017}
}

@inproceedings{gidaris2018rotation,
title={Unsupervised Representation Learning by Predicting Image Rotations},
author={Spyros Gidaris and Praveer Singh and Nikos Komodakis},
booktitle={International Conference on Learning Representations},
year={2018},
url={https://openreview.net/forum?id=S1v4N2l0-},
}

@inproceedings{hadsell2006dimensionality,
title={Dimensionality reduction by learning an invariant mapping},
author={Hadsell, Raia and Chopra, Sumit and LeCun, Yann},
booktitle={Proceedings of the IEEE/CVF conference on computer vision and pattern recognition},
volume={2},
pages={1735--1742},
year={2006},
organization={IEEE}
}

@InProceedings{wu2018unsupervised,
author = {Wu, Zhirong and Xiong, Yuanjun and Yu, Stella X. and Lin, Dahua},
title = {Unsupervised Feature Learning via Non-Parametric Instance Discrimination},
booktitle = {Proceedings of the IEEE Conference on Computer Vision and Pattern Recognition},
month = {June},
year = {2018}
}

@article{oord2018cpc,
title={Representation learning with contrastive predictive coding},
author={Oord, Aaron van den and Li, Yazhe and Vinyals, Oriol},
journal={arXiv preprint arXiv:1807.03748},
year={2018}
}

@inproceedings{hjelm2018dim,
title={Learning deep representations by mutual information estimation and maximization},
author={R Devon Hjelm and Alex Fedorov and Samuel Lavoie-Marchildon and Karan Grewal and Phil Bachman and Adam Trischler and Yoshua Bengio},
booktitle={International Conference on Learning Representations},
year={2019},
url={https://openreview.net/forum?id=Bklr3j0cKX},
}

@article{bromley1993siamese,
title={Signature verification using a" siamese" time delay neural network},
author={Bromley, Jane and Guyon, Isabelle and LeCun, Yann and S{\"a}ckinger, Eduard and Shah, Roopak},
journal={Advances in neural information processing systems},
volume={6},
year={1993}
}

@inproceedings{he2020moco,
title={Momentum contrast for unsupervised visual representation learning},
author={He, Kaiming and Fan, Haoqi and Wu, Yuxin and Xie, Saining and Girshick, Ross},
booktitle={Proceedings of the IEEE/CVF conference on computer vision and pattern recognition},
pages={9729--9738},
year={2020}
}

@inproceedings{chen2021mocov3,
title={An empirical study of training self-supervised vision transformers},
author={Chen, Xinlei and Xie, Saining and He, Kaiming},
booktitle={Proceedings of the IEEE/CVF international conference on computer vision},
pages={9640--9649},
year={2021}
}

@inproceedings{chen2020simsiam,
title={A simple framework for contrastive learning of visual representations},
author={Chen, Ting and Kornblith, Simon and Norouzi, Mohammad and Hinton, Geoffrey},
booktitle={International conference on machine learning},
pages={1597--1607},
year={2020},
organization={PmLR}
}

@inproceedings{chen2020simclr,
title={A simple framework for contrastive learning of visual representations},
author={Chen, Ting and Kornblith, Simon and Norouzi, Mohammad and Hinton, Geoffrey},
booktitle={International conference on machine learning},
pages={1597--1607},
year={2020},
organization={PmLR}
}

@article{grill2020byol,
  title={Bootstrap your own latent-a new approach to self-supervised learning},
  author={Grill, Jean-Bastien and Strub, Florian and Altch{\'e}, Florent and Tallec, Corentin and Richemond, Pierre and Buchatskaya, Elena and Doersch, Carl and Avila Pires, Bernardo and Guo, Zhaohan and Gheshlaghi Azar, Mohammad and others},
  journal={Advances in neural information processing systems},
  volume={33},
  pages={21271--21284},
  year={2020}
}

@InProceedings{caron2021dino,
    author    = {Caron, Mathilde and Touvron, Hugo and Misra, Ishan and J\'egou, Herv\'e and Mairal, Julien and Bojanowski, Piotr and Joulin, Armand},
    title     = {Emerging Properties in Self-Supervised Vision Transformers},
    booktitle = {Proceedings of the IEEE/CVF International Conference on Computer Vision (ICCV)},
    month     = {October},
    year      = {2021},
    pages     = {9650-9660}
}

@inproceedings{zhou2022ibot,
title={Image {BERT} Pre-training with Online Tokenizer},
author={Jinghao Zhou and Chen Wei and Huiyu Wang and Wei Shen and Cihang Xie and Alan Yuille and Tao Kong},
booktitle={International Conference on Learning Representations},
year={2022},
url={https://openreview.net/forum?id=ydopy-e6Dg}
}

@inproceedings{vincent2008dae,
title={Extracting and composing robust features with denoising autoencoders},
author={Vincent, Pascal and Larochelle, Hugo and Bengio, Yoshua and Manzagol, Pierre-Antoine},
booktitle={Proceedings of the 25th international conference on Machine learning},
pages={1096--1103},
year={2008}
}

@inproceedings{pathak2016context,
  title={Context encoders: Feature learning by inpainting},
  author={Pathak, Deepak and Krahenbuhl, Philipp and Donahue, Jeff and Darrell, Trevor and Efros, Alexei A},
  booktitle={Proceedings of the IEEE conference on computer vision and pattern recognition},
  pages={2536--2544},
  year={2016}
}

@inproceedings{he2022mae,
  title={Masked autoencoders are scalable vision learners},
  author={He, Kaiming and Chen, Xinlei and Xie, Saining and Li, Yanghao and Doll{\'a}r, Piotr and Girshick, Ross},
  booktitle={Proceedings of the IEEE/CVF conference on computer vision and pattern recognition},
  pages={16000--16009},
  year={2022}
}

@inproceedings{bao2022beit,
title={{BE}iT: {BERT} Pre-Training of Image Transformers},
author={Hangbo Bao and Li Dong and Songhao Piao and Furu Wei},
booktitle={International Conference on Learning Representations},
year={2022},
url={https://openreview.net/forum?id=p-BhZSz59o4}
}

@inproceedings{assran2023jepa,
title={Self-supervised learning from images with a joint-embedding predictive architecture},
author={Assran, Mahmoud and Duval, Quentin and Misra, Ishan and Bojanowski, Piotr and Vincent, Pascal and Rabbat, Michael and LeCun, Yann and Ballas, Nicolas},
booktitle={Proceedings of the IEEE/CVF Conference on Computer Vision and Pattern Recognition},
pages={15619--15629},
year={2023}
}

@article{radford2018gpt,
title={Improving language understanding by generative pre-training},
author={Radford, Alec and Narasimhan, Karthik and Salimans, Tim and Sutskever, Ilya and others},
journal={OpenAI blog},
year={2018},
publisher={San Francisco, CA, USA}
}

@article{radford2019gpt2,
title={Language models are unsupervised multitask learners},
author={Radford, Alec and Wu, Jeffrey and Child, Rewon and Luan, David and Amodei, Dario and Sutskever, Ilya and others},
journal={OpenAI blog},
volume={1},
number={8},
pages={9},
year={2019}
}

@article{brown2020gpt3,
title={Language models are few-shot learners},
author={Brown, Tom and Mann, Benjamin and Ryder, Nick and Subbiah, Melanie and Kaplan, Jared D and Dhariwal, Prafulla and Neelakantan, Arvind and Shyam, Pranav and Sastry, Girish and Askell, Amanda and others},
journal={Advances in neural information processing systems},
volume={33},
pages={1877--1901},
year={2020}
}

@article{achiam2023gpt4,
title={Gpt-4 technical report},
author={Achiam, Josh and Adler, Steven and Agarwal, Sandhini and Ahmad, Lama and Akkaya, Ilge and Aleman, Florencia Leoni and Almeida, Diogo and Altenschmidt, Janko and Altman, Sam and Anadkat, Shyamal and others},
journal={arXiv preprint arXiv:2303.08774},
year={2023}
}

@inproceedings{chen2020igpt,
title={Generative pretraining from pixels},
author={Chen, Mark and Radford, Alec and Child, Rewon and Wu, Jeffrey and Jun, Heewoo and Luan, David and Sutskever, Ilya},
booktitle={International conference on machine learning},
pages={1691--1703},
year={2020},
organization={PMLR}
}

@inproceedings{esser2021vqgan,
title={Taming transformers for high-resolution image synthesis},
author={Esser, Patrick and Rombach, Robin and Ommer, Bjorn},
booktitle={Proceedings of the IEEE/CVF conference on computer vision and pattern recognition},
pages={12873--12883},
year={2021}
}

@inproceedings{ramesh2021dalle,
title={Zero-shot text-to-image generation},
author={Ramesh, Aditya and Pavlov, Mikhail and Goh, Gabriel and Gray, Scott and Voss, Chelsea and Radford, Alec and Chen, Mark and Sutskever, Ilya},
booktitle={International conference on machine learning},
pages={8821--8831},
year={2021},
organization={Pmlr}
}

@inproceedings{lee2022rqvae,
  title={Autoregressive image generation using residual quantization},
  author={Lee, Doyup and Kim, Chiheon and Kim, Saehoon and Cho, Minsu and Han, Wook-Shin},
  booktitle={Proceedings of the IEEE/CVF conference on computer vision and pattern recognition},
  pages={11523--11532},
  year={2022}
}

@article{tian2024var,
title={Visual autoregressive modeling: Scalable image generation via next-scale prediction},
author={Tian, Keyu and Jiang, Yi and Yuan, Zehuan and Peng, Bingyue and Wang, Liwei},
journal={Advances in neural information processing systems},
volume={37},
pages={84839--84865},
year={2024}
}

@article{sun2024llamagen,
title={Autoregressive model beats diffusion: Llama for scalable image generation},
author={Sun, Peize and Jiang, Yi and Chen, Shoufa and Zhang, Shilong and Peng, Bingyue and Luo, Ping and Yuan, Zehuan},
journal={arXiv preprint arXiv:2406.06525},
year={2024}
}

@article{fan2024fluid,
title={Fluid: Scaling autoregressive text-to-image generative models with continuous tokens},
author={Fan, Lijie and Li, Tianhong and Qin, Siyang and Li, Yuanzhen and Sun, Chen and Rubinstein, Michael and Sun, Deqing and He, Kaiming and Tian, Yonglong},
journal={arXiv preprint arXiv:2410.13863},
year={2024}
}

@article{li2024mar,
  title={Autoregressive image generation without vector quantization},
  author={Li, Tianhong and Tian, Yonglong and Li, He and Deng, Mingyang and He, Kaiming},
  journal={Advances in Neural Information Processing Systems},
  volume={37},
  pages={56424--56445},
  year={2024}
}

@article{wu2025dcar,
  title={Dc-ar: Efficient masked autoregressive image generation with deep compression hybrid tokenizer},
  author={Wu, Yecheng and Chen, Junyu and Zhang, Zhuoyang and Xie, Enze and Yu, Jincheng and Chen, Junsong and Hu, Jinyi and Lu, Yao and Han, Song and Cai, Han},
  journal={arXiv preprint arXiv:2507.04947},
  year={2025}
}

@inproceedings{chang2022maskgit,
title={Maskgit: Masked generative image transformer},
author={Chang, Huiwen and Zhang, Han and Jiang, Lu and Liu, Ce and Freeman, William T},
booktitle={Proceedings of the IEEE/CVF conference on computer vision and pattern recognition},
pages={11315--11325},
year={2022}
}

@inproceedings{li2023mage,
  title={Mage: Masked generative encoder to unify representation learning and image synthesis},
  author={Li, Tianhong and Chang, Huiwen and Mishra, Shlok and Zhang, Han and Katabi, Dina and Krishnan, Dilip},
  booktitle={Proceedings of the IEEE/CVF Conference on Computer Vision and Pattern Recognition},
  pages={2142--2152},
  year={2023}
}

@article{pang2024randar,
title={RandAR: Decoder-only Autoregressive Visual Generation in Random Orders},
author={Pang, Ziqi and Zhang, Tianyuan and Luan, Fujun and Man, Yunze and Tan, Hao and Zhang, Kai and Freeman, William T. and Wang, Yu-Xiong},
journal={arXiv preprint arXiv:2412.01827},
year={2024}
}

@article{van2017vqvae,
  title={Neural discrete representation learning},
  author={Van Den Oord, Aaron and Vinyals, Oriol and others},
  journal={Advances in neural information processing systems},
  volume={30},
  year={2017}
}

@article{rao1999predictive,
title={Predictive coding in the visual cortex: a functional interpretation of some extra-classical receptive-field effects},
author={Rao, Rajesh PN and Ballard, Dana H},
journal={Nature neuroscience},
volume={2},
number={1},
pages={79--87},
year={1999},
publisher={Nature Publishing Group}
}

@inproceedings{radford2021clip,
title={Learning transferable visual models from natural language supervision},
author={Radford, Alec and Kim, Jong Wook and Hallacy, Chris and Ramesh, Aditya and Goh, Gabriel and Agarwal, Sandhini and Sastry, Girish and Askell, Amanda and Mishkin, Pamela and Clark, Jack and others},
booktitle={International conference on machine learning},
pages={8748--8763},
year={2021},
organization={PmLR}
}

@article{dosovitskiy2020vit,
  title={An image is worth 16x16 words: Transformers for image recognition at scale},
  author={Dosovitskiy, Alexey and Beyer, Lucas and Kolesnikov, Alexander and Weissenborn, Dirk and Zhai, Xiaohua and Unterthiner, Thomas and Dehghani, Mostafa and Minderer, Matthias and Heigold, Georg and Gelly, Sylvain and others},
  journal={arXiv preprint arXiv:2010.11929},
  year={2020}
}

@article{oquab2023dinov2,
title={Dinov2: Learning robust visual features without supervision},
author={Oquab, Maxime and Darcet, Timoth{\'e}e and Moutakanni, Th{\'e}o and Vo, Huy and Szafraniec, Marc and Khalidov, Vasil and Fernandez, Pierre and Haziza, Daniel and Massa, Francisco and El-Nouby, Alaaeldin and others},
journal={arXiv preprint arXiv:2304.07193},
year={2023}
}

@article{chu2024visionllama,
title={Visionllama: A unified llama interface for vision tasks},
author={Chu, Xiangxiang and Su, Jianlin and Zhang, Bo and Shen, Chunhua},
journal={CoRR},
year={2024}
}

@misc{siméoni2025dinov3,
title={DINOv3}, 
author={Oriane Siméoni and Huy V. Vo and Maximilian Seitzer and Federico Baldassarre and Maxime Oquab and Cijo Jose and Vasil Khalidov and Marc Szafraniec and Seungeun Yi and Michaël Ramamonjisoa and Francisco Massa and Daniel Haziza and Luca Wehrstedt and Jianyuan Wang and Timothée Darcet and Théo Moutakanni and Leonel Sentana and Claire Roberts and Andrea Vedaldi and Jamie Tolan and John Brandt and Camille Couprie and Julien Mairal and Hervé Jégou and Patrick Labatut and Piotr Bojanowski},
year={2025},
eprint={2508.10104},
archivePrefix={arXiv},
primaryClass={cs.CV},
url={https://arxiv.org/abs/2508.10104}, 
}

@inproceedings{touvron2021cait,
title={Going deeper with image transformers},
author={Touvron, Hugo and Cord, Matthieu and Sablayrolles, Alexandre and Synnaeve, Gabriel and J{\'e}gou, Herv{\'e}},
booktitle={Proceedings of the IEEE/CVF international conference on computer vision},
pages={32--42},
year={2021}
}

@article{su2024roformer,
title={Roformer: Enhanced transformer with rotary position embedding},
author={Su, Jianlin and Ahmed, Murtadha and Lu, Yu and Pan, Shengfeng and Bo, Wen and Liu, Yunfeng},
journal={Neurocomputing},
volume={568},
pages={127063},
year={2024},
publisher={Elsevier}
}

@misc{shazeer2020glu,
title={GLU Variants Improve Transformer}, 
author={Noam Shazeer},
year={2020},
eprint={2002.05202},
archivePrefix={arXiv},
primaryClass={cs.LG},
url={https://arxiv.org/abs/2002.05202}, 
}

@misc{
hendrycks2017gelu,
title={Bridging Nonlinearities and Stochastic Regularizers with Gaussian Error Linear Units},
author={Dan Hendrycks and Kevin Gimpel},
year={2017},
url={https://openreview.net/forum?id=Bk0MRI5lg}
}

@inproceedings{henry2020qknorm,
title = "Query-Key Normalization for Transformers",
author = "Henry, Alex  and
      Dachapally, Prudhvi Raj  and
      Pawar, Shubham Shantaram  and
      Chen, Yuxuan",
editor = "Cohn, Trevor  and
      He, Yulan  and
      Liu, Yang",
booktitle = "Findings of the Association for Computational Linguistics: EMNLP 2020",
month = nov,
year = "2020",
address = "Online",
publisher = "Association for Computational Linguistics",
url = "https://aclanthology.org/2020.findings-emnlp.379/",
doi = "10.18653/v1/2020.findings-emnlp.379",
pages = "4246--4253",
}

@InProceedings{dehghani2023vit22b,
title = 	 {Scaling Vision Transformers to 22 Billion Parameters},
author =       {Dehghani, Mostafa and Djolonga, Josip and Mustafa, Basil and Padlewski, Piotr and Heek, Jonathan and Gilmer, Justin and Steiner, Andreas Peter and Caron, Mathilde and Geirhos, Robert and Alabdulmohsin, Ibrahim and Jenatton, Rodolphe and Beyer, Lucas and Tschannen, Michael and Arnab, Anurag and Wang, Xiao and Riquelme Ruiz, Carlos and Minderer, Matthias and Puigcerver, Joan and Evci, Utku and Kumar, Manoj and Steenkiste, Sjoerd Van and Elsayed, Gamaleldin Fathy and Mahendran, Aravindh and Yu, Fisher and Oliver, Avital and Huot, Fantine and Bastings, Jasmijn and Collier, Mark and Gritsenko, Alexey A. and Birodkar, Vighnesh and Vasconcelos, Cristina Nader and Tay, Yi and Mensink, Thomas and Kolesnikov, Alexander and Pavetic, Filip and Tran, Dustin and Kipf, Thomas and Lucic, Mario and Zhai, Xiaohua and Keysers, Daniel and Harmsen, Jeremiah J. and Houlsby, Neil},
booktitle = 	 {Proceedings of the 40th International Conference on Machine Learning},
pages = 	 {7480--7512},
year = 	 {2023},
editor = 	 {Krause, Andreas and Brunskill, Emma and Cho, Kyunghyun and Engelhardt, Barbara and Sabato, Sivan and Scarlett, Jonathan},
volume = 	 {202},
series = 	 {Proceedings of Machine Learning Research},
month = 	 {23--29 Jul},
publisher =    {PMLR},
url = 	 {https://proceedings.mlr.press/v202/dehghani23a.html},
}

@inproceedings{esser2024sd3,
title={Scaling Rectified Flow Transformers for High-Resolution Image Synthesis},
author={Patrick Esser and Sumith Kulal and Andreas Blattmann and Rahim Entezari and Jonas M{\"u}ller and Harry Saini and Yam Levi and Dominik Lorenz and Axel Sauer and Frederic Boesel and Dustin Podell and Tim Dockhorn and Zion English and Robin Rombach},
booktitle={Forty-first International Conference on Machine Learning},
year={2024},
url={https://openreview.net/forum?id=FPnUhsQJ5B}
}

@misc{ba2016layernormalization,
title={Layer Normalization}, 
author={Jimmy Lei Ba and Jamie Ryan Kiros and Geoffrey E. Hinton},
year={2016},
eprint={1607.06450},
archivePrefix={arXiv},
primaryClass={stat.ML},
url={https://arxiv.org/abs/1607.06450}, 
}

@InProceedings{xiao2018upernet,
author = {Xiao, Tete and Liu, Yingcheng and Zhou, Bolei and Jiang, Yuning and Sun, Jian},
title = {Unified Perceptual Parsing for Scene Understanding},
booktitle = {Proceedings of the European Conference on Computer Vision (ECCV)},
month = {September},
year = {2018}
}

@misc{touvron2023llama,
title={LLaMA: Open and Efficient Foundation Language Models}, 
author={Hugo Touvron and Thibaut Lavril and Gautier Izacard and Xavier Martinet and Marie-Anne Lachaux and Timothée Lacroix and Baptiste Rozière and Naman Goyal and Eric Hambro and Faisal Azhar and Aurelien Rodriguez and Armand Joulin and Edouard Grave and Guillaume Lample},
year={2023},
eprint={2302.13971},
archivePrefix={arXiv},
primaryClass={cs.CL},
url={https://arxiv.org/abs/2302.13971}, 
}

@misc{bai2023qwentechnicalreport,
title={Qwen Technical Report}, 
author={Jinze Bai and Shuai Bai and Yunfei Chu and Zeyu Cui and Kai Dang and Xiaodong Deng and Yang Fan and Wenbin Ge and Yu Han and Fei Huang and Binyuan Hui and Luo Ji and Mei Li and Junyang Lin and Runji Lin and Dayiheng Liu and Gao Liu and Chengqiang Lu and Keming Lu and Jianxin Ma and Rui Men and Xingzhang Ren and Xuancheng Ren and Chuanqi Tan and Sinan Tan and Jianhong Tu and Peng Wang and Shijie Wang and Wei Wang and Shengguang Wu and Benfeng Xu and Jin Xu and An Yang and Hao Yang and Jian Yang and Shusheng Yang and Yang Yao and Bowen Yu and Hongyi Yuan and Zheng Yuan and Jianwei Zhang and Xingxuan Zhang and Yichang Zhang and Zhenru Zhang and Chang Zhou and Jingren Zhou and Xiaohuan Zhou and Tianhang Zhu},
year={2023},
eprint={2309.16609},
archivePrefix={arXiv},
primaryClass={cs.CL},
url={https://arxiv.org/abs/2309.16609}, 
}

@inproceedings{loshchilov2018adamw,
title={Decoupled Weight Decay Regularization},
author={Ilya Loshchilov and Frank Hutter},
booktitle={International Conference on Learning Representations},
year={2019},
url={https://openreview.net/forum?id=Bkg6RiCqY7},
}

@misc{goyal2018warmup,
title={Accurate, Large Minibatch SGD: Training ImageNet in 1 Hour}, 
author={Priya Goyal and Piotr Dollár and Ross Girshick and Pieter Noordhuis and Lukasz Wesolowski and Aapo Kyrola and Andrew Tulloch and Yangqing Jia and Kaiming He},
year={2018},
eprint={1706.02677},
archivePrefix={arXiv},
primaryClass={cs.CV},
url={https://arxiv.org/abs/1706.02677}, 
}

@InProceedings{touvron2021deit,
title = 	 {Training data-efficient image transformers and distillation through attention},
author =       {Touvron, Hugo and Cord, Matthieu and Douze, Matthijs and Massa, Francisco and Sablayrolles, Alexandre and Jegou, Herve},
booktitle = 	 {Proceedings of the 38th International Conference on Machine Learning},
pages = 	 {10347--10357},
year = 	 {2021},
editor = 	 {Meila, Marina and Zhang, Tong},
volume = 	 {139},
series = 	 {Proceedings of Machine Learning Research},
month = 	 {18--24 Jul},
publisher =    {PMLR},
url = 	 {https://proceedings.mlr.press/v139/touvron21a.html},
}

@misc{clark2020llrd,
title={ELECTRA: Pre-training Text Encoders as Discriminators Rather Than Generators}, 
author={Kevin Clark and Minh-Thang Luong and Quoc V. Le and Christopher D. Manning},
year={2020},
eprint={2003.10555},
archivePrefix={arXiv},
primaryClass={cs.CL},
url={https://arxiv.org/abs/2003.10555}, 
}

@inproceedings{dubuk2020randaugement,
author = {Cubuk, Ekin Dogus and Zoph, Barret and Shlens, Jon and Le, Quoc},
booktitle = {Advances in Neural Information Processing Systems},
editor = {H. Larochelle and M. Ranzato and R. Hadsell and M.F. Balcan and H. Lin},
pages = {18613--18624},
publisher = {Curran Associates, Inc.},
title = {RandAugment: Practical Automated Data Augmentation with a Reduced Search Space},
url = {https://proceedings.neurips.cc/paper_files/paper/2020/file/d85b63ef0ccb114d0a3bb7b7d808028f-Paper.pdf},
volume = {33},
year = {2020}
}

@INPROCEEDINGS{szegedy2016label,
  author={Szegedy, Christian and Vanhoucke, Vincent and Ioffe, Sergey and Shlens, Jon and Wojna, Zbigniew},
  booktitle={2016 IEEE Conference on Computer Vision and Pattern Recognition (CVPR)}, 
  title={Rethinking the Inception Architecture for Computer Vision}, 
  year={2016},
  volume={},
  number={},
  pages={2818-2826},
  doi={10.1109/CVPR.2016.308}}

@inproceedings{zhang2018mixup,
title={mixup: Beyond Empirical Risk Minimization},
author={Hongyi Zhang and Moustapha Cisse and Yann N. Dauphin and David Lopez-Paz},
booktitle={International Conference on Learning Representations},
year={2018},
url={https://openreview.net/forum?id=r1Ddp1-Rb},
}

@INPROCEEDINGS{yun2019cutmix,
author={Yun, Sangdoo and Han, Dongyoon and Chun, Sanghyuk and Oh, Seong Joon and Yoo, Youngjoon and Choe, Junsuk},
booktitle={2019 IEEE/CVF International Conference on Computer Vision (ICCV)}, 
title={CutMix: Regularization Strategy to Train Strong Classifiers With Localizable Features}, 
year={2019},
volume={},
number={},
pages={6022-6031},
doi={10.1109/ICCV.2019.00612}}

@InProceedings{huang2016droppath,
author="Huang, Gao
and Sun, Yu
and Liu, Zhuang
and Sedra, Daniel
and Weinberger, Kilian Q.",
editor="Leibe, Bastian
and Matas, Jiri
and Sebe, Nicu
and Welling, Max",
title="Deep Networks with Stochastic Depth",
booktitle="Computer Vision -- ECCV 2016",
year="2016",
publisher="Springer International Publishing",
address="Cham",
pages="646--661",
isbn="978-3-319-46493-0"
}

@InProceedings{zhou2017ade20k,
author = {Zhou, Bolei and Zhao, Hang and Puig, Xavier and Fidler, Sanja and Barriuso, Adela and Torralba, Antonio},
title = {Scene Parsing Through ADE20K Dataset},
booktitle = {Proceedings of the IEEE Conference on Computer Vision and Pattern Recognition (CVPR)},
month = {July},
year = {2017}
}

@article{russakovsky2015imagenet1k,
Author = {Olga Russakovsky and Jia Deng and Hao Su and Jonathan Krause and Sanjeev Satheesh and Sean Ma and Zhiheng Huang and Andrej Karpathy and Aditya Khosla and Michael Bernstein and Alexander C. Berg and Li Fei-Fei},
Title = { {ImageNet Large Scale Visual Recognition Challenge} },
Year = {2015},
journal   = {International Journal of Computer Vision (IJCV)},
doi = {10.1007/s11263-015-0816-y},
volume={115},
number={3},
pages={211-252}
}

@misc{mmseg2020,
    title={{MMSegmentation}: OpenMMLab Semantic Segmentation Toolbox and Benchmark},
    author={MMSegmentation Contributors},
    howpublished = {\url{https://github.com/open-mmlab/mmsegmentation}},
    year={2020}
}

@article{goodfellow2020gan,
  title={Generative adversarial networks},
  author={Goodfellow, Ian and Pouget-Abadie, Jean and Mirza, Mehdi and Xu, Bing and Warde-Farley, David and Ozair, Sherjil and Courville, Aaron and Bengio, Yoshua},
  journal={Communications of the ACM},
  volume={63},
  number={11},
  pages={139--144},
  year={2020},
  publisher={ACM New York, NY, USA}
}

@article{ho2020ddpm,
  title={Denoising diffusion probabilistic models},
  author={Ho, Jonathan and Jain, Ajay and Abbeel, Pieter},
  journal={Advances in neural information processing systems},
  volume={33},
  pages={6840--6851},
  year={2020}
}

@inproceedings{song2021ddim,
  title={Denoising Diffusion Implicit Models},
  author={Song, Jiaming and Meng, Chenlin and Ermon, Stefano},
  booktitle={International Conference on Learning Representations}
}

@inproceedings{peebles2023dit,
  title={Scalable diffusion models with transformers},
  author={Peebles, William and Xie, Saining},
  booktitle={Proceedings of the IEEE/CVF international conference on computer vision},
  pages={4195--4205},
  year={2023}
}

@article{tarvainen2017mean,
  title={Mean teachers are better role models: Weight-averaged consistency targets improve semi-supervised deep learning results},
  author={Tarvainen, Antti and Valpola, Harri},
  journal={Advances in neural information processing systems},
  volume={30},
  year={2017}
}

@misc{rw2019timm,
  author = {Ross Wightman},
  title = {PyTorch Image Models},
  year = {2019},
  publisher = {GitHub},
  journal = {GitHub repository},
  doi = {10.5281/zenodo.4414861},
  howpublished = {\url{https://github.com/rwightman/pytorch-image-models}}
}

@inproceedings{wolf-etal-2020-transformers,
    title = "Transformers: State-of-the-Art Natural Language Processing",
    author = "Thomas Wolf and Lysandre Debut and Victor Sanh and Julien Chaumond and Clement Delangue and Anthony Moi and Pierric Cistac and Tim Rault and Rémi Louf and Morgan Funtowicz and Joe Davison and Sam Shleifer and Patrick von Platen and Clara Ma and Yacine Jernite and Julien Plu and Canwen Xu and Teven Le Scao and Sylvain Gugger and Mariama Drame and Quentin Lhoest and Alexander M. Rush",
    booktitle = "Proceedings of the 2020 Conference on Empirical Methods in Natural Language Processing: System Demonstrations",
    month = oct,
    year = "2020",
    address = "Online",
    publisher = "Association for Computational Linguistics",
    url = "https://www.aclweb.org/anthology/2020.emnlp-demos.6",
    pages = "38--45"
}

@inproceedings{lhoest-etal-2021-datasets,
    title = "Datasets: A Community Library for Natural Language Processing",
    author = "Lhoest, Quentin  and
      Villanova del Moral, Albert  and
      Jernite, Yacine  and
      Thakur, Abhishek  and
      von Platen, Patrick  and
      Patil, Suraj  and
      Chaumond, Julien  and
      Drame, Mariama  and
      Plu, Julien  and
      Tunstall, Lewis  and
      Davison, Joe  and
      {\v{S}}a{\v{s}}ko, Mario  and
      Chhablani, Gunjan  and
      Malik, Bhavitvya  and
      Brandeis, Simon  and
      Le Scao, Teven  and
      Sanh, Victor  and
      Xu, Canwen  and
      Patry, Nicolas  and
      McMillan-Major, Angelina  and
      Schmid, Philipp  and
      Gugger, Sylvain  and
      Delangue, Cl{\'e}ment  and
      Matussi{\`e}re, Th{\'e}o  and
      Debut, Lysandre  and
      Bekman, Stas  and
      Cistac, Pierric  and
      Goehringer, Thibault  and
      Mustar, Victor  and
      Lagunas, Fran{\c{c}}ois  and
      Rush, Alexander  and
      Wolf, Thomas",
    booktitle = "Proceedings of the 2021 Conference on Empirical Methods in Natural Language Processing: System Demonstrations",
    month = nov,
    year = "2021",
    address = "Online and Punta Cana, Dominican Republic",
    publisher = "Association for Computational Linguistics",
    url = "https://aclanthology.org/2021.emnlp-demo.21",
    pages = "175--184",
    eprint={2109.02846},
    archivePrefix={arXiv},
    primaryClass={cs.CL},
}

@article{taesiri2023imagenet,
  title={Imagenet-hard: The hardest images remaining from a study of the power of zoom and spatial biases in image classification},
  author={Taesiri, Mohammad Reza and Nguyen, Giang and Habchi, Sarra and Bezemer, Cor-Paul and Nguyen, Anh},
  journal={Advances in Neural Information Processing Systems},
  volume={36},
  pages={35878--35953},
  year={2023}
}

@inproceedings{neuhaus2023spurious,
  title={Spurious features everywhere-large-scale detection of harmful spurious features in imagenet},
  author={Neuhaus, Yannic and Augustin, Maximilian and Boreiko, Valentyn and Hein, Matthias},
  booktitle={Proceedings of the IEEE/CVF International Conference on Computer Vision},
  pages={20235--20246},
  year={2023}
}

@inproceedings{el2024scalable,
  title={Scalable pre-training of large autoregressive image models},
  author={El-Nouby, Alaaeldin and Klein, Michal and Zhai, Shuangfei and Bautista, Miguel Angel and Shankar, Vaishaal and Toshev, Alexander and Susskind, Joshua M and Joulin, Armand},
  booktitle={Proceedings of the 41st International Conference on Machine Learning},
  pages={12371--12384},
  year={2024}
}

@inproceedings{fini2025multimodal,
  title={Multimodal autoregressive pre-training of large vision encoders},
  author={Fini, Enrico and Shukor, Mustafa and Li, Xiujun and Dufter, Philipp and Klein, Michal and Haldimann, David and Aitharaju, Sai and da Costa, Victor G Turrisi and B{\'e}thune, Louis and Gan, Zhe and others},
  booktitle={Proceedings of the Computer Vision and Pattern Recognition Conference},
  pages={9641--9654},
  year={2025}
}
}

\end{document}